\documentclass{llncs}

\usepackage{makeidx}  
\usepackage{amsmath}
\usepackage{amssymb}
\usepackage{times}
\usepackage{verbatim}
\usepackage{algorithmic}
\usepackage[ruled,vlined]{algorithm2e}

\usepackage{graphicx}
\DeclareGraphicsExtensions{.pdf}
\usepackage{booktabs}
\usepackage{multicol}
\usepackage{multirow}

\usepackage{color}

\begin{document}
\frontmatter          
\pagestyle{headings}  
\addtocmark{Local Optima Networks} 
\mainmatter              
\title{Local Optima Networks: A New Model of Combinatorial Fitness Landscapes}
\titlerunning{Local Optima Networks}  

\author{Gabriela Ochoa\inst{1} \and S\'ebastien Verel\inst{2} \and Fabio Daolio\inst{3} \and Marco Tomassini\inst{3}}
\tocauthor{Gabriela Ochoa, S\'ebastien Verel, Fabio Daolio, Marco Tomassini}
\institute{Computing Science and Mathematics, University of Stirling, Scotland.\and INRIA Lille - Nord Europe and University of Nice Sophia-Antipolis, France.
\and Information Systems Department, University of Lausanne, Switzerland.}

\maketitle              

\begin{abstract}
This chapter overviews a recently introduced network-based model of combinatorial landscapes: Local Optima Networks (LON). The model compresses the information given by the whole search space into a smaller mathematical object that is a graph having as vertices the local optima and as edges the possible weighted transitions between them. Two definitions of edges have been proposed: basin-transition and escape-edges, which capture relevant topological features of the underlying search spaces. This network model brings a new set of metrics to characterize the structure of combinatorial landscapes, those associated with the science of complex networks. These metrics are described, and results are presented of local optima network extraction and analysis for two selected combinatorial landscapes: NK landscapes and the quadratic assignment problem. Network features are found to correlate with and even predict the performance of heuristic search algorithms operating on these problems.

\keywords{fitness landscapes, complex networks, combinatorial optimization, local optima, basin of attraction, NK landscapes, Quadratic Assignment Problem }
\end{abstract}

\section{Introduction}
The fitness landscape metaphor appears most commonly when describing the dynamics of evolutionary algorithms, and its origins are attributed to the population geneticist Sewall Wright \cite{Wright:32}. However, the metaphor can be used for computational search in general; the search space can be regarded as a spacial structure where each point (candidate solution) has a height (objective function value) forming a landscape surface. In this scenario, the search process would be an adaptive-walk over a landscape that can range from having many peaks of high fitness boarding  deep cliffs to valleys of low fitness, to being smooth, with low hills and gentle valleys.

Identifying the landscape features affecting the effectiveness of heuristic search algorithms is relevant for both predicting their performance and improving their design. Some properties of landscapes that are known to have a strong influence on heuristic search are the number of local optima or peaks in the landscape, the distribution of the local optima in the search space, the correlation between fitness values of neighboring points in the landscape, the topology of the basins of attraction of the local optima, and the presence of neutrality (different search points having the same objective value). Statistical methods have been proposed to measure some of these properties, for example, fitness-distance correlation~\cite{jones95b}, distributions of solutions density~\cite{ROS:96}, landscape correlation functions~\cite{weinberger90}, and the negative slope coefficient~\cite{nsc06}.  These metrics work by sampling the landscape surface to provide an approximation of its shape. An alternative view, first  introduced in chemical physics in the study  of energy landscapes~\cite{stillinger95}, is to construct a network formed by the landscape local optima  (minima or maxima). In this view of energy surfaces, the network's vertices are energy minima and there is an edge between two minima if the system can jump from one to the other with an energy cost of the order of the thermal energies. Usually this ``transition state'' goes through a low energy barrier such as a saddle point in the surface. The resulting graph has been referred to as \textit{inherent network}. Recent work by Doye and coworkers and by Caflish and coworkers~\cite{doye02,doye05,rao-caflisch04} has  shown the benefits of this approach:  it provides a synthetic view of the energy landscape and the network can be studied using appropriate statistical methods to characterize it in various ways~\cite{newman03}. For example, Doye et al.~\cite{doye02,doye05} found that the inherent networks of the energy landscapes of small atomic clusters are often of the scale-free type with a power-law degree distribution function, featuring a kind of single or multiple ``funnel'' structure. The global energy minimum is the most highly connected node at the bottom of the funnel. This means that the path to the global energy minimum is easy to follow starting anywhere in the energy landscape. The concept of community structure of a network, introduced first for social networks~\cite{newman03}, has also been applied, showing that in some cases energy minima split into almost separate groups or communities~\cite{doye05-comm}. This effect is even more spectacular for polypeptides~\cite{gfeller-07}. This kind of information is invaluable for understanding the dynamics induced on the energy landscape such as cluster rearrangements or protein folding.

The {\em local optima networks} fitness landscape model, described in this chapter, adapts the notion of the inherent network of energy surfaces to the realm of combinatorial (discrete) search spaces. As for energy surfaces (which exist in continuous space), the vertices correspond to solutions that are minima or maxima of the associated combinatorial problem, but edges are defined differently. The combinatorial counterpart considers oriented and weighted edges. In a first version,  the weights represent an approximation to the probability of transition between the respective basins in a given direction \cite{gecco08,ppsn10,pre08,alife08,tec11}. This definition, although informative, produced densely connected networks and required  exhaustive sampling of the basins of attraction. A second version,  {\em escape edges} was proposed in \cite{ea11}, which  does not require a full computation of the basins. Instead, these edges  account for the chances of escaping a local optimum after a controlled mutation (e.g. 1 or 2 bit-flips in binary space) followed by hill-climbing.  As a first benchmark case in the study of local optima networks,  the well studied family of abstract landscapes, the Kauffman's $NK$ model, was selected \cite{Kauffman1987,kauffman93}. In this model the ruggedness, and hence the difficulty of the landscape, can be tuned from easy to hard. Two NK models incorporating {\em neutrality} (i.e. extended regions of equal or quasi-equal fitness) were considered: the $NK_p$ (`probabilistic' $NK$)~\cite{barnett98}, and $NK_q$ (`quantized' $NK$)~\cite{newman98} families. Subsequently, a more complex and realistic search space was studied. Specifically, the quadratic assignment problem (QAP) introduced by Koopmans and Beckmann~\cite{Koopmans57}, which is known to be NP-hard~\cite{sahni76}.

The local optima network model captures in detail the number and distribution of local optima in the search space; features which are known to be of utmost importance for understanding the search difficulty of the corresponding landscape. This understanding may be exploited when designing efficient search algorithms. For example, it has been observed in many combinatorial landscapes that local optima are not randomly distributed, rather they tend to be clustered in a ``central massif" (or ``big valley" if we are minimizing). This globally convex landscape structure has been observed in the $NK$ family of landscapes \cite{Kauffman1987,kauffman93}, and in many combinatorial optimization problems, such as the traveling salesman problem \cite{boese94}, graph bipartitioning~\cite{merz98}, and flowshop scheduling \cite{reeves99}. Search algorithms exploiting this global structure have been proposed~\cite{boese94,reeves99}. For the travelling salesman problem, the big-valley structure holds in much of the search space. However,  it has been recently found that the big-valley structure disappears, giving rise to multiple funnels, around local optima that are very close to the global optimum \cite{HainsWH11}. A specialized crossover operator has been proposed to exploit and overcome this multi-funnel structure \cite{WhitleyHH10}.

The analysis of local optima networks so far has shown interesting correlations between network features and known search difficulty on the studied combinatorial problems. This chapter overviews the conception and analysis of local optima networks. A brief account of the science of complex networks is given before describing the combinatorial landscapes, relevant definitions and methods employed. A summary of the most relevant results of the analysis is presented, and finally, the prospects of this research effort are discussed.

\section{The Science of Complex Networks}

\label{cn}

\index{Complex Networks}

The last few years have seen an increased interest in the structure of the big networks that form part of our daily environment such as the World Wide Web, the Internet, transportation and electrical power networks, web-based social networks such as Facebook, and many others. These networks have properties that are unparalleled in simple graphs such as lattices, properties that are akin  to those of \textit{complex systems} in general. In these systems, it is difficult or even impossible to infer global behaviors given the rules that are obeyed by the system components and their interactions. For this reason, these big networks are called \textit{complex networks} and  their structure gives rise to a wide range of
dynamical behaviors. Since this chapter draws heavily on complex network nomenclature and methods, to make it self-contained to a large extent, we give a brief introduction to the field. There exist many references on complex networks: a technical but still very readable introductory book is~\cite{dorogotsev}, while~\cite{newman-book} is a comprehensive reference.

Mathematically, networks are just \textit{graphs} $G(V,E)$ where $V$ is the set of \textit{vertices} and $E$ is the set of edges that join pairs of vertices. A complex network class that enjoys a precise mathematical description is \textit{random graphs} which are
introduced below. Random graphs are a useful abstraction that can sometimes be used to model real networks or, at least, to compare with actual complex networks.

\subsection{Random Graphs}
\label{rand-graphs}

The random-graph model was formally defined by Erd\"os and R\'enyi at the end of the 1950s. In its simplest form, the model consists of $N$ vertices joined by edges that are placed between pairs of vertices uniformly at random. In other words, each of the possible $N(N-1)/2$ edges is present with probability $p$ and absent with probability $1-p$. The model is  often referred to as  $G_{N,p}$ to point out that, rigorously speaking, there is no such thing as a random graph, but rather an ensemble $G_{N,p}$ of equiprobable graphs.

Another closely related model of a random graph considers the family of graphs $G_{N,M}$ with $N$ vertices and exactly $M$ edges. For $0 \le M \le {N \choose 2}$, there are $s = {N(N-1)/2\choose M}$ graphs with $M$ edges. If the probability of selecting any one of them is $1/s$, then the ensemble $G_{N,M}$ is called the family of uniform random graphs. For $M \simeq pN$, the two models are very similar, but we shall use $G_{N,p}$ in what follows.

A few simple facts are worth noting about random graphs. The \textit{average degree} $\bar k $ of a graph $G$ is the average of all the vertex degrees in $G$: $\bar k  = (1/N) \sum _{j=1}^N k_j$, where $k_j$ is the degree of vertex $j$. If $|E| = M$ is the number of edges in $G$, then $M = (N  \bar k )/ 2$, since $\sum _{j=1}^N k_j=2M$ (each edge is counted twice).

The expected number of edges of a random graph belonging to $G_{N,p}$ is clearly $(1/2) N(N-1) p$, but since each edge has two ends, the average number of edge ends is $ N(N-1) p$, which in turn means that the average degree of a vertex in a random graph is

\begin{equation}
\bar k=\frac {N(N-1) p}{N} = (N-1) p \simeq Np
\label{eq:av-degr-rg}
\end{equation}

\noindent for sufficiently large $N$.

An important property of a connected random graph is that the average path length, i.e the mean distance between nodes, is of the order of $\log N$, which means that any two nodes are only a short distance apart since $\log N$ grows very slowly with increasing $N$.

\subsection{Other Network Topologies}
\label{swn}

Random graphs are interesting objects as they obey, in a probabilistic sense, general mathematical properties.  They are also a useful model for generating problem instances for testing network algorithms, and they are used in other ways too. But are random graphs a useful model of the networks that permeate society? Actually, social scientists felt qualitatively as early as the 1950s that social and professional links and acquaintances did not follow a random structure.  For example, if a person has some relationship with two others, then the latter two are more likely to know each other
than are two arbitrary persons. This does not fit the random-graph model, however, where the likelihood that two given nodes are connected is the same independent of any other consideration. In a ground-breaking paper, Watts and Strogatz~\cite{watts-strogatz-98} proposed a simple network
construction algorithm that gives rise to graphs having the following properties: the path length from any node to any other node is short, as in random graphs; but, unlike random graphs, there is local structure in the network. Watts and Strogatz called their networks \textit{small-world networks}, a term that has been in use for a long time in the field of social games to indicate that there is a small separation between any two persons in a large social network.

The discovery of these new properties was made possible by the abundance of online network data and the computer power to analyze these data; something that was not available to social scientists at earlier times.  Many networks have been studied since, both man-made and natural: the Internet, the World Wide Web, scientific collaboration and coauthorship networks, metabolic and neural networks, air traffic, telephone calls, e-mail exchanges, and many others~\cite{newman-book}. Most of these studies have confirmed that, indeed, real networks are not random in the sense of random-graph theory, and they possess a number of quite interesting properties.

Some definitions of global and local network properties that will be used in the rest of the chapter are described below.

\subsection{Some Graph Statistics}
\label{graph-stat}

Drawing and visualizing a network with up to a few tens of nodes may help in understanding its structure. However, when there are thousands of nodes,
this is no longer possible. For this reason, a number of statistics have been proposed to describe the main features of a graph. Taken together, these statistics characterize the nature of a network.

Four statistics are particularly useful: the average degree, already defined in Sect. \ref{rand-graphs}, the \textit{clustering coefficient}, the  \textit{average path length}, and the \textit{degree distribution function}. We shall now briefly describe these graph measures. A fuller treatment can be found in~\cite{newman-book}.

\paragraph{Clustering Coefficient.}
Here we use the following definition of clustering: consider a particular node $j$ in a graph, and let us assume that it has degree $k$, i.e. it has $k$ edges connecting it to its $k$ neighboring nodes. If all $k$ vertices in the neighborhood were completely connected to each other, forming a  clique, then the number of edges would be equal to ${k} \choose{2}$. The clustering coefficient $C_j$ of node $j$  is defined as
the ratio between the $e$ edges that actually exist between the $k$ neighbors and the number of possible edges between these nodes

\begin{equation}
C_j = \frac{e}{{{k} \choose{2}}} = \frac{2e}{k(k-1)}
\label{eq:clust-coeff}
\end{equation}

\noindent Thus $C_j$ is a measure of the ``cliquishness'' of a neighborhood: the higher the value of $C_j$, the more likely it is that two vertices that are adjacent to a third one are also neighbors of each other.

\begin{figure} [!ht]
  \begin{center}
    \includegraphics[width=\textwidth]{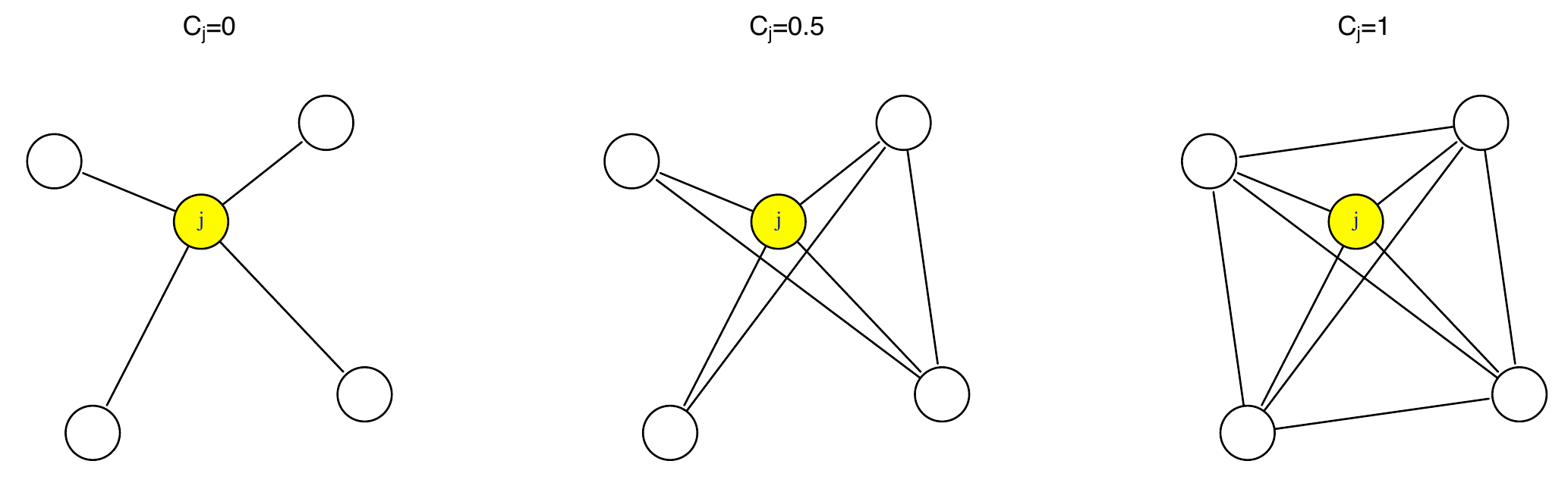} \protect \\
  \end{center}
  \caption{In the left image the clustering coefficient of node $j$, $C_j=0$ since there are no links
  among $j$'s neighbors. In the middle image $C_j=0.5$ because three out of the possible six
  edges among the neighbors of $j$ are present. In the right image $C_j=1$ as all the edges that
  could be there are actually present (it is a clique).
\label{cc}}
\end{figure}

For example, in Fig.~\ref{cc}, the leftmost case has $C_j = 0$ since none of the links between  node $j$'s neighbors is present. In the middle figure, three out of the possible six links are present and thus $C_j = (2 \times 3) / (4 \times 3) = 6/12 = 0.5$, while in the rightmost
case $C_j = 1$ since all six links between $j$'s neighbors are present.

The \textit{average clustering coefficient} $\bar C $ is the average of $C_i$ over all $N$ vertices
$i \in V(G)$: $\bar C  = (1/N )\sum _{i=1}^N C_i.$ The clustering coefficient of a graph $G$ thus expresses the degree of locality of the connections.

The clustering coefficient of a random graph is simply $\bar k / N \simeq p = \bar C $, where \textit{N} is the total number of vertices and $p$ is the probability that there is an edge between any two vertices since all edges are equiprobable and uncorrelated. One thus sees that the clustering coefficient of a random graph decreases with the graph size $N$ and approaches $0$ for $N \rightarrow \infty$. The clustering coefficient of a complete graph is $1$, since each of a node's neighbors are connected to each other by definition.

\paragraph{Average Path Length.}
The \textit{distance} between two nodes $i$ and $j$ is defined as the number of edges between $i$ and $j$. We denote the shortest path  between nodes $i,j \in V(G)$ by $l_{ij}$ as being the path with the shortest distance. The average, or mean, path length $\bar L $ of $G$ is then defined as

\begin{equation}
\bar L  =   \frac{2}{N(N-1)}\sum _{i=1}^N \sum _{j > i} l_{ij}
\label{eq:av-path-len}
\end{equation}

\noindent The normalizing constant $2/N(N-1)$ is the inverse of the total number of pairs of vertices. If there is no path between any two nodes, their distance is conventionally set to $\infty$ (note that Eq.~\ref{eq:av-path-len} does not hold in this case).

The mean path length gives an idea of ``how long'' it takes to navigate a connected network. Random graphs and small-world networks share the property that $\bar L$ scales as $\log N$ and thus most vertices in these networks are connected by a short path. This is not the case in $d$-dimensional regular lattice graphs, where $\bar L $ scales as $N^{1/d}$. For instance, in a ring $\bar L $ scales linearly with $N$ and is inversely proportional to $k$, the number of neighbors.

\paragraph{Degree Distribution Function.}

The degree distribution $P(k)$ of an undirected graph $G$ is a function that gives the probability that a randomly selected vertex has degree $k$. $P(k)$ can also be seen as the fraction of vertices in the graph that have degree $k$. Similar definitions also apply for the in-links
and out-links of the vertices in a directed graph for which one can define a degree distribution function for both the outgoing  $P_{out}(k)$ and the  the incoming $P_{in}(k)$ links.

For a random graph with connection probability $p$, the probability $P(k)$ that a random node has degree $k$ is given by
\begin{equation}
P(k) = {{N-1} \choose {k}} p^k (1-p)^{N-1-k}
\label{eq:deg-distr-fun}
\end{equation}

This is the number of ways in which $k$ edges can be selected from a certain node out of the $N-1$ possible edges, given that the edges can be chosen independently of each other and have the same probability $p$. Thus $P(k)$ is a binomial distribution peaked at $P(\bar k )\simeq Np$, as already found in Equation (\ref{eq:av-degr-rg}). Since this distribution has a  rapidly diminishing tail, most nodes will have similar degrees. Low- and high-degree nodes, say a few standard deviations away from the mean, have a negligible probability, since the tails fall off very rapidly. Networks having this degree distribution will thus be rather homogeneous as far as the connectivity is concerned. For large $N$ and for $pN$ constant, the binomial
distribution can be well approximated by  the Poisson distribution of mean $\bar k = Np$:
\begin{equation}
P(k) = e^{-\bar k}\: \frac{{\bar k}^k} {k!}
\label{eq:deg-distr-fun-poiss}
\end{equation}\\

\noindent Another rapidly-decaying degree distribution function that appears in model graphs  is the exponential distribution:
\begin{equation}
P(k) \propto e^{-k/\bar k}
\label{eq:deg-distr-fun-exp}
\end{equation}
\noindent This degree distribution results when nodes are progressively added to a growing network such that a new node has the same probability of forming a link with any of the already existing nodes. Most real networks, however, do not show this kind of behavior.  Instead, the so called  \textit{scale-free networks}, a model of which was first proposed by
Barab\'asis and Albert~\cite{barabasi99}, seem to be closer to real life networks. In these networks, $P(k)$ follows a power-law distribution:
\begin{equation}
P(k) = c\, k^{-\gamma},
\label{eq:deg-distr-fun-power}
\end{equation}
\noindent  where $c$ and
$\gamma$ are positive constants.

In scale-free networks, while most nodes have a low degree, there is a small but non-negligible number of highly connected nodes. This structure has a profound influence on the dynamics of processes taking place on those networks. It is worth mentioning that this model has been recently criticized \cite{Clauset2009} as it turns out that, upon close inspection, many empirical data-sets in the literature that were previously assumed to have a power-low distribution are better modeled by alternative distributions.

Poisson, exponential, and power-law distributions are characteristic of model random and scale-free graphs respectively. The empirical distribution functions found for real-life graphs are seldom of this type though, because it is almost impossible to find such ``pure'' networks among finite
sampled ones. However, most real networks have degree distributions  that are fat-tailed, i.e. the right part of the distribution extends to regions that would have negligible probability for a Poisson distribution; in other words, nodes with high degree exist with non-negligible probability. Two distributions that have been useful to fit real data are the \textit{power-law with exponential cutoff} and the \textit{stretched exponential}. Both
forms take into account that in a finite network there must be a maximum finite degree. As an example, the following is an exponentially-truncated power-law:
\begin{equation}
P(k) \propto k^{-\gamma} \, \exp(-k/k_c),
\label{eq:deg-distr-fun-trunc}
\end{equation}
\noindent where $k_c$ is a ``critical connectivity". When $k$ approaches $k_c$ the exponential
term tends to $0$ and $P(k)$ decreases faster than a power-law due to the exponential cutoff.

To conclude, we can say that the degree distribution function, together with the other statistics, are a kind of rough ``signature'' of the type of network and can be helpful in predicting the main aspects of the properties of the network. The reader should be aware that there are other measures beyond those described here, such as those that say which are the most ``central'' actors. Likewise, the picture has been one of static networks: their dynamical properties, have been neglected here.  A good source for advanced material is~\cite{newman-book}.

\subsection{Weighted Networks}
\label{WN}

Weighted networks are a useful extension of the network model. Weights $w(e)$ are assigned to edges $e \in E$ and could represent, for instance, the bandwidth of a communication line, the number of passengers transported on a given air route, the distance between two metro stations, and many other real-life aspects of networks. Here, weights will represent transition probabilities among optima and their basins in fitness landscapes. We denote $w_{ij}$ as the transition probability between local optima $i$ and $j$, which in our model is generally different than the transition from $j$ to $i$, denoted by $w_{ji}$ (see Section \ref{edges} for more details).

Statistics for weighted networks are more or less straightforward extensions of those used for unweighted networks. Those that will be used in the rest of the chapter are briefly outlined below. The reader is refereed to~\cite{bart05} for more details.

Suitable distribution functions can also be defined for weighted networks. For example, it can be of interest to know the function $P(w_{e})$ which indicates the frequency of weight $w$ among the edges $e$ in a given weighted network. Since our networks are directed, we use $P_{in}(w_{e})$ and $P_{out}(w_{e})$ which give the probability that any given edge $e$ has incoming or outgoing weight $w$.

\paragraph{Strength.}
The vertex \textit{strength}, $s_i$, is defined as $s_i = \sum_{j \in {\cal V}(i) - \{i\}}  w_{ij}  $, where the sum is over the set ${\cal V}(i) - \{i\}$ of neighbors of vertex $i$. This metric is, therefore, a generalization of the node's degree giving information about the number and importance of the edges.

\paragraph{Weighted Clustering Coefficient.}

The standard  clustering coefficient (described in Section \ref{graph-stat}) does not consider weighted edges. We thus use the {\em weighted clustering} measure proposed by~\cite{bart05}, which combines the topological information with the weight distribution of the network:

$$c^{w}(i) = \frac{1}{s_i(k_i - 1)} \sum_{j,h} \frac{w_{ij} + w_{ih}}{2} a_{ij} a_{jh} a_{hi}$$
where $s_i$ is the vertex strength,  $s_i = \sum_{j \in {\cal V}(i) - \{i\}}  w_{ij}  $, $a_{nm} = 1$ if $w_{nm} > 0$,
$a_{nm} = 0$ if $w_{nm} = 0$ and $k_i = \sum_{j \not= i} a_{ij}$.

This metric $c^{w}(i)$ counts, for each triple formed in the neighborhood of vertex $i$ (indicated in the equation by $a_{ij} a_{jh} a_{hi}$), the weight of the two participating edges of vertex $i$. The normalization factor $s_i(k_i - 1)$, ensures that the metric is in the range $[0,1]$. It is customary to define $C^w$ as the weighted clustering coefficient averaged over all vertices of the network.

\paragraph{Disparity.}

The {\em disparity} measure  $Y_{2}(i)$,  gauges the heterogeneity of the contributions of the edges of node $i$ to the total weight (strength):

$$Y_{2}(i) = \sum_{j \not= i} \left( \frac{w_{ij}}{s_i} \right)^2 $$

\paragraph{Shortest Paths.}

For weighted graphs, computing the shortest paths depends on the meaning of the edge weights. For example, if the weight represents e.g. a frequency of interaction, an electrical load, the number of passengers transported or a probability of transition, then the higher the weight, the ``nearer''  the two end points. In this case the path length between any two connected vertices is taken as the sum of the reciprocal of the weights $\sum 1/w_{ij}$ where the sum is over all edges $\{ij\}$ traversed along the path from the start node to the end node. On the other hand, if weights represent ``costs'' of some kind and the aim is to have low total cost, then the length is simply the sum of the costs of all edges along the path and the minimal length is that of the path with minimum cost.

In the local optima network model, we measure the shortest distance between two nodes as the expected number of operator moves to go from one node to the other. Given that the transition probability between two nodes $i$ and $j$ is given by $w_{ij}$, we calculate the distance between them as $d_{ij} = 1 / w_{ij}$. The length of a path between two arbitrary connected nodes is, therefore, the sum of these distances along the edges connecting them. The average path length of the whole network is the average value of all the possible shortest paths.

\subsection{Community Structure in Networks}
\label{sec:community}

\index{Community Structure}

A last theme in this section that we want to treat briefly is the ``intermediate'' structure of large networks since it will play a role in the following and is an important feature of complex networks. Model networks grown according to the Barab\'asi--Albert recipe \cite{barabasi99} or randomly generated have little structure in the sense that there are few or no recognizable sub-networks. That is, if one looks at a picture of the network it appears to be rather homogeneous on a global scale.

On the contrary, many observed networks, especially those arising from social interactions, show the presence of clusters of nodes. These clusters are called \textit{communities}. It is difficult, if not impossible, to give a precise and unique mathematical definition of a community. An intuitive definition of a community is the following: nodes belonging to a community are more strongly associated with each other than they are with the rest of the network. In other words, the intra-community connectivity is higher than the inter-community connectivity.  Of course, the definition is somewhat circular but in the last few years several algorithms have been proposed for community detection. Since this task is a hard computational
problem, machine learning algorithms and heuristics have been used and in practice these work satisfactorily.

\section{Example Combinatorial Landscapes}
\label{landscapes}
\subsection{The NK model}
\index{NK Landscapes} 

The idea of an $NK$ landscape is to have $N$ ``spins``or ``loci'', each with two possible values, $0$ or $1$. The model is a real stochastic function $\Phi$ defined on binary strings $s \in  \{0,1\}^N$ of length $N$, $\Phi: s \rightarrow \mathbb{R_+}$. The value of $K$ determines how many other gene values in the string influence a given gene  $s_i, \: i=1,\ldots,N$. The value of $\Phi$ is the average of the contributions $\phi_i$ of all the loci:

$$
\Phi(s) = \frac{1} {N} \sum_{i=1}^N \phi_i(s_i, s_{{i}_1}, \ldots, s_{{i}_K})
$$

By increasing the value of $K$ from 0 to $N-1$, $NK$ landscapes can be tuned from smooth to rugged. For $K=0$ all contributions can be optimized independently which makes $\Phi$ a simple additive function with a single maximum. At the other extreme, when $K=N-1$, the landscape becomes completely random. The probability of any given configuration being the optimum is $1/(N+1)$, and the expected number of local optima is $2^N/(N+1)$. Intermediate values of $K$ interpolate between these two extremes and have a variable degree of ``epistasis'', i.e. of  gene interaction~\cite{kauffman93,kaul06,limic04}.

The $K$ variables that form the context of the fitness contribution of gene $s_i$ can be chosen according to different models. The two most widely studied models are the {\em random neighborhood} model, where the $K$  variables are chosen randomly according to a uniform distribution among the $N-1$ variables other than $s_i$, and the {\em adjacent neighborhood} model, in which the $K$ variables are closest to $s_i$ in a total ordering $s_1, s_2, \ldots, s_N$ (using periodic boundaries). No significant differences between the two models were found in terms of  global properties of the respective families of landscapes, such as mean number of local optima or autocorrelation length \cite{weinberger91,kauffman93}. Similarly, our
preliminary studies on the characteristics of the $NK$ landscape optima networks did not show noticeable differences between the two neighborhood models. Therefore, the study in this chapter considers the more general random model.

\subsection{The Quadratic Assignment Problem}
\label{sec:qapdef}

\index{Quadratic Assignment Problem} 
\index{QAP}

The Quadratic Assignment Problem (QAP)  is a combinatorial problem in which a set of facilities with given flows has to be assigned to a set of locations with given distances in such a way that the sum of the product of flows and distances is minimized. A solution to the QAP is generally written as a permutation $\pi$ of the set $\{1,2,...,n\}$. The cost associated with a permutation $\pi$ is given by:

$$
 C(\pi)=\sum_{i=1}^{n}\sum_{j=1}^{n}{a_{ij}b_{\pi_{i}\pi_{j}}}
$$

\noindent where $n$ denotes the number of facilities/locations and  $A=\{a_{ij}\}$ and $B=\{b_{ij}\}$ are referred to as the distance and flow matrices, respectively. The structure of these two matrices  characterizes the class of instances of the QAP problem.

The results presented in  this chapter are based on two instance generators proposed in~\cite{Knowles2003emo}. These generators were originally devised for the multi-objective QAP, but were adapted for the single-objective QAP and used for the local optima network analysis in \cite{daolio2011communities,cec10}. In order to perform a statistical analysis of the extracted local optima networks, several problem instances of the two different problem classes were considered. The first generator produces uniformly random instances where all flows and distances are integers sampled from uniform distributions.  This leads to the kind of problem known in literature as \emph{Tai}\verb"nn"\emph{a}, being \verb"nn" the problem dimension~\cite{Taillard1995}. The second generator produces flow entries that are non-uniform random values. This procedure, detailed in~\cite{Knowles2003emo} generates random instances of type \emph{Tai}\verb"nn"\emph{b} which have the so called ``real-like'' structure since they resemble the structure of QAP problems found in practical applications.

\section{The Local Optima Network Model}
\label{lon}

This section formally describes the the local optima network model of combinatorial landscapes. We start by defining the notion of fitness landscapes,  and follow by formalizing the notions of nodes and edges of the network model.

A fitness landscape~\cite{reidys2002combinatorial} is a triplet $(S, V, f)$ where $S$ is a set of potential solutions i.e. a search space; $V : S \longrightarrow 2^S$, a neighborhood structure, is a function that assigns to every $s \in S$ a set of neighbors $V(s)$, and $f : S \longrightarrow R$ is a fitness function that can be pictured as the \textit{height} of the corresponding solutions.

Local optima networks have been analyzed for the two combinatorial landscapes discussed in section~\ref{landscapes}. Therefore, two search spaces or solution representations have been studied: binary strings ($NK$ landscapes) and permutations (QAP). For each case, the most basic neighborhood structure is considered, as described in Table \ref{tab:moves}. The single bit-flip operation changes a single bit in a given binary string, whereas the pairwise exchange operation exchanges any two positions in a permutation, thus transforming it into another permutation.

\begin{table}[!ht]
\begin{center}
 \caption{Search space and neighborhood structure characteristics. \label{tab:moves}}
\begin{tabular}{lcccc}
\toprule
  {\em Representation} & {\em Length} & {\em Search space size} & {\em Neighborhood} & {\em Neighborhood size} \\
\midrule
 Binary & $N$ & $2^N$ & single bit-flip &  $N$ \\
 Permutation & $N$ & $N!$ & pairwise exchange & $N(N-1)/2$ \\
\bottomrule
\end{tabular}
\end{center}
\end{table}

\subsection{Definition of Nodes}

We start by describing the $HillClimbing$ algorithm (Algorithm~\ref{algoHC}) used to determine the local optima, and therefore define the basins of attraction. The algorithm defines a mapping from the search space $S$ to the set of locally optimal solutions $S^*$. Hill climbing algorithms differ in their so-called {\em pivot} or selection rule. In best-improvement local search, the entire neighborhood is explored and the best solution is returned, whereas in first-improvement, a neighbor is selected uniformly at random and is accepted if it improves on the current fitness value. We consider here a best-improvement local search heuristic (see  Algorithm~\ref{algoHC}). For a comparison between first and best-improvement local optima network models, the reader is referred to~\cite{ppsn10}.

\index{Hill Climbing}

\begin{algorithm}[t]
\begin{algorithmic}
\STATE Choose initial solution $s \in S$
\REPEAT
    \STATE choose $s^{'} \in V(s)$, such that $f(s^{'}) = max_{x \in V(s)} f(x)$
        \IF{$f(s) < f(s^{'})$}
            \STATE $s \leftarrow s^{'}$
    \ENDIF
\UNTIL{$s$ is a  Local optimum}
 \end{algorithmic}
\caption{Best-improvement local search (hill-climbing).}
\label{algoHC}
\end{algorithm}

This best-improvement local search (or hill-climbing) algorithm  is used to determine the local optima. The neighborhoods used for each of the studied representation can be seen in Table \ref{tab:moves}. These local optima will represent the nodes of the network as discussed below.

\paragraph{Nodes.} A local optimum ($LO$), which is taken to be a maximum here, is a solution $s^{*}$ such that $\forall s \in V(s)$, $f(s) \leq f(s^{*})$.

\index{Local Optima}
\index{Local Optimum}

Let us denote by $h(s)$ the stochastic operator that associates each solution $s$ to its local optimum, i.e. the solution obtained after applying  the  best-improvement  hill-climbing algorithm (see Algorithm~\ref{algoHC}) until convergence. The size of the landscape is finite, so we can denote by $LO_1$, $LO_2$, $LO_3 \ldots, LO_p$, the local optima. These $LOs$ are the vertices of the \textit{local optima network}.

\subsection{Definition of Edges}
\label{edges}

Two edge models have been considered: basin-transition and escape edges.

\paragraph{Basin-transition edges.}

The basin of attraction of a local optimum $LO_i \in S$ is the set $b_i = \{s \in S ~|~ h(s) = LO_i \}$. The size of the basin of attraction of a local optimum $i$ is the cardinality of $b_i$, denoted $\sharp b_i$. Notice that for non-neutral\footnote{For a definition of basins that deals with neutrality, the reader is referred to~\cite{tec11}.}
 fitness landscapes, as are standard $NK$ landscapes, the basins of attraction as defined above produce a partition of the configuration space $S$. Therefore, $S = \cup_{i \in S^{*}} b_i$ and $\forall i \in S$ $\forall j \not= i$, $b_i \cap b_j = \emptyset$.

We can now define the weight of an edge that connects two feasible solutions in the  fitness landscape.

\noindent For each pair of solutions $s$ and $s^{'}$, $p(s \rightarrow s^{'} )$ is the probability to pass from $s$ to $s^{'}$ with the given neighborhood structure. These probabilities are given below for the two solution representations studied (see Table \ref{tab:moves}), with length $N$ or size $N$ and considering uniform selection of random neighbors.

\begin{description}

\item {Binary representation:}
\index{Binary Strings}

if $s^{'} \in V(s)$ , $p(s \rightarrow s^{'} ) = \frac{1}{N}$ and \\
\indent if $s^{'} \not\in V(s)$ , $p(s \rightarrow s^{'} ) = 0$.

\item{Permutation representation:}
\index{Permutation}

\noindent if $s^{'} \in V(s)$ , $p(s \rightarrow s^{'} ) = \frac{1}{N(N-1)/2}$ and \\
\indent if $s^{'} \not\in V(s)$ , $p(s \rightarrow s^{'} ) = 0$.

\end{description}

\noindent The probability ( $p(s \rightarrow b_j ) \leq 1$) to go from solution $s
\in S$ to a solution belonging to the basin $b_j$, is:
$$
p(s \rightarrow b_j ) = \sum_{s^{'} \in b_j} p(s \rightarrow s^{'} ) \quad
$$

\noindent Thus, the total probability of going from basin $b_i$ to basin $b_j$, i.e. the weight $w_{ij}$ of edge $e_{ij}$, is the average over all $s \in b_i$ of the transition probabilities  to solutions $s^{'} \in b_j$ :

$$p(b_i \rightarrow b_j) = \frac{1}{\sharp b_i} \sum_{s \in b_i} p(s \rightarrow b_j ) \quad $$

\paragraph{Escape edges.}

The escape edges are defined according to a distance function $d$ (minimal number of moves between two solutions), and a positive integer $D > 0$. There is an edge $e_{ij}$ between $LO_i$ and $LO_j$ if a solution $s$ exists such that $d(s, LO_i) \leq D$ and $s$ $h(s) = LO_j$. The weight $w_{ij}$ of this edge is $w_{ij}= \sharp \{ s \in S ~|~ d(s, LO_i) \leq D \mbox{ and } h(s) = LO_j \}$. This weight can be normalized by the number of solutions,  $\sharp \{ s \in S ~|~ d(s, LO_i) \leq D\}$, within reach at distance $D$.

\subsection{Local optima network}

The weighted local optima network $G_w=(N,E)$ is the graph where the nodes $n_i \in N$ are the local optima, and there is an edge $e_{ij} \in E$, with  weight $w_{ij}$, between two nodes $n_i$ and $n_j$ if $w_{ij} > 0$.

According to both definitions of edge weights, $w_{ij}$ may be different than $w_{ji}$. Thus, two weights are needed in general, and we have an oriented transition graph.

\begin{figure}[htb!]
\begin{center}
\includegraphics[width=0.7\textwidth]{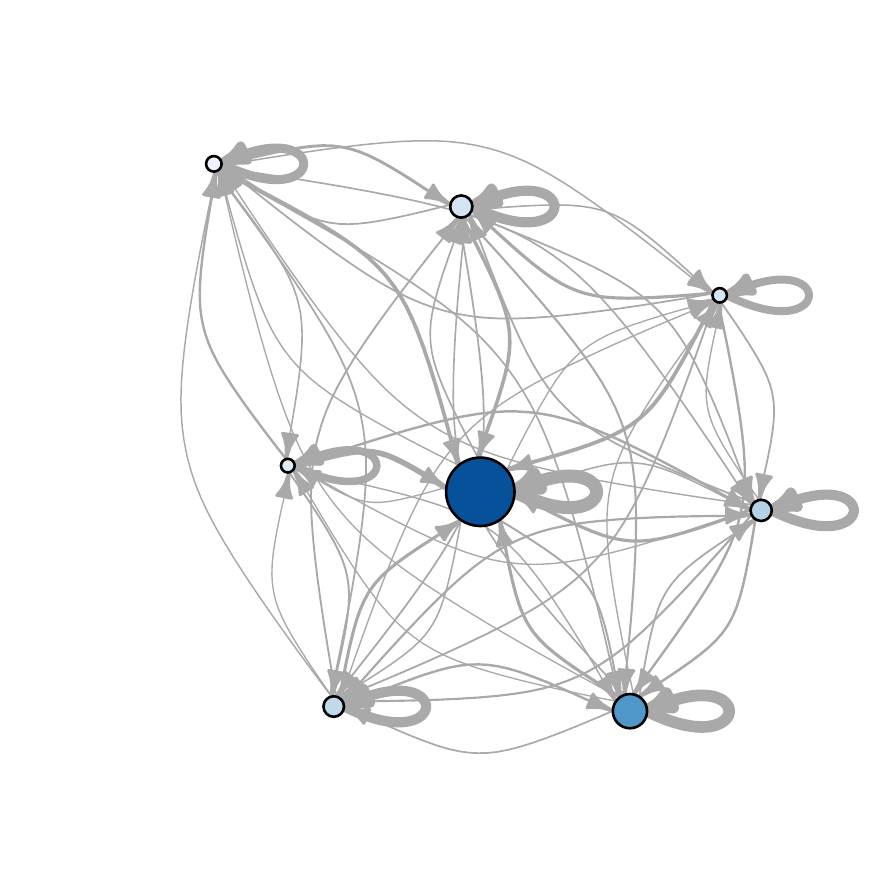}
\caption{\small Local optima network with basin-transition edges for an $NK$-landscape instance with N = 18, K = 2. The size of the nodes is proportional to the basin sizes. The nodes' color represent the fitness values: the darker the color, the highest the fitness value. The edges' width scales with the transition probability (weight)  between local optima. \label{fig:lonb} }
\label{fig:lons}
\end{center}
\end{figure}

\begin{figure} [!ht]
\begin{center}
\begin{tabular}{cc}
($D=1$) & ($D=2$) \\
\includegraphics[width=0.52\textwidth]{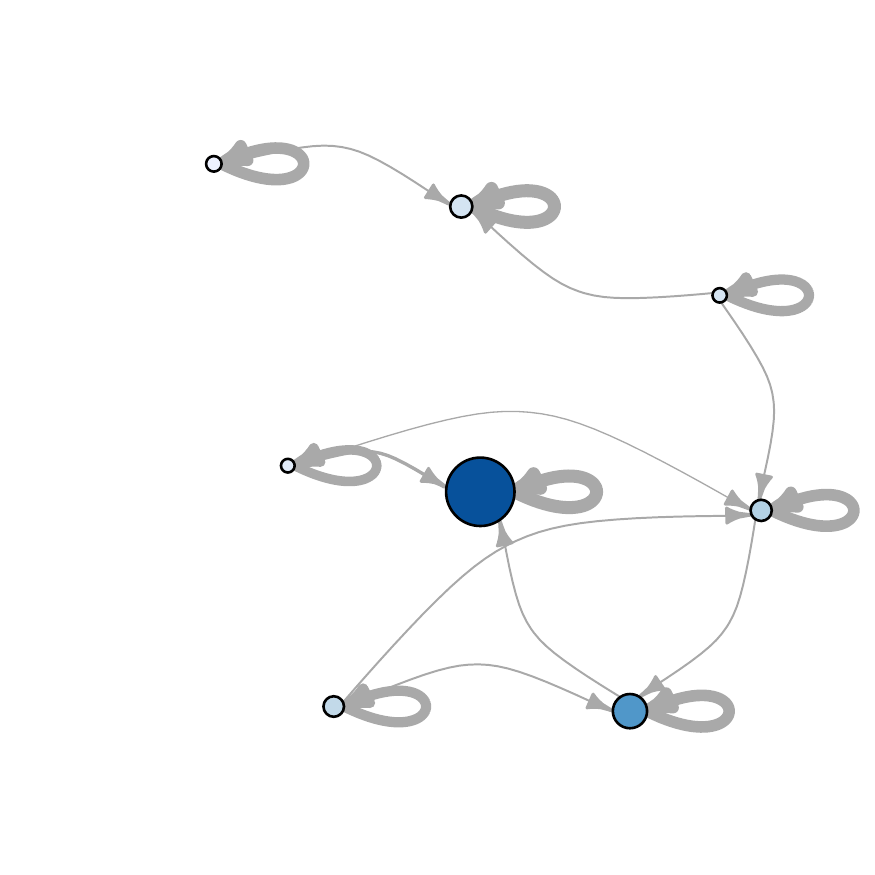} &
\includegraphics[width=0.52\textwidth]{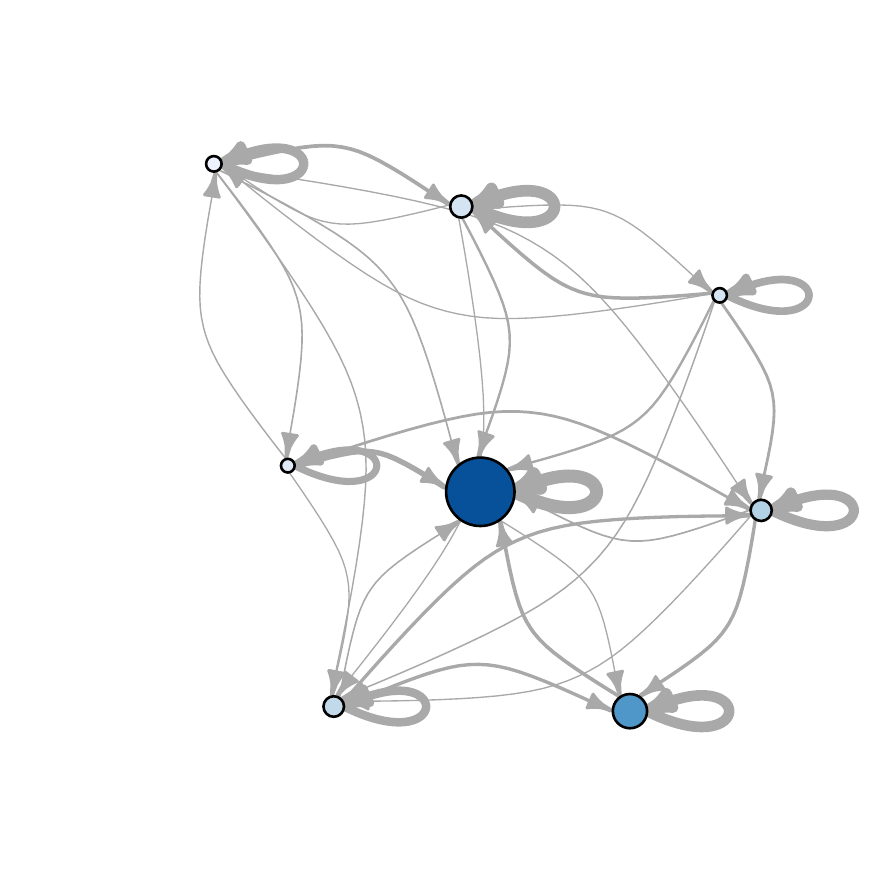} \\
\end{tabular}
\end{center}
\caption{\small Local optima network with escape edges (with $D=1$ and $D=2$) for an $NK$-landscape instance with N = 18, K = 2. Since this model does not require the calculation of the basins sizes, these are not depicted in the plots.  The nodes' color represent the fitness values: the darker the color, the highest the fitness value. The edges' width scales with the transition probability (weight) between local optima.\label{fig:lone}}
\end{figure}

Figures \ref{fig:lonb} and \ref{fig:lone} illustrate the alternative local optima network (LON) models. All figures correspond to a real $NK$ landscape with N = 18, K = 2. Figure \ref{fig:lonb} illustrates the basin-transition edges, while Figure \ref{fig:lone} the escape edges with  $D=1$  (left) and $D=2$ (right). Notice that the basin-transition edges (Figure \ref{fig:lonb}) produce a densely connected network, while the escape edges (Figure \ref{fig:lone})  produce more sparse networks.

\section{Results of the Network Analysis}

The purpose of this section is to give an overview of the main results of the analysis of local optima networks for the two example combinatorial landscapes: $NK$ landscapes (Section \ref{ResultsNK}) and the Quadratic Assignment Problem (Section \ref{ResultsQAP}). For each example, the empirical set up and instances analyzed are discussed. The values obtained from the study of basins of attraction, general network metrics and connectivity, are reported and discussed.

\subsection{The NK model}
\label{ResultsNK}

\index{NK Landscape}
For the $NK$ model, the two definitions of edges, i.e. basin-transition and escape edges (section \ref{edges}), are considered. Moreover, an initial study correlating network metrics with search difficulty is also presented.

Results are presented for landscapes with $N = 18$ and varied values of $K$ ($K \in \{2, 4, 6, 8, 10, 12, 14, 16, 17 \}$. $N=18$ represents the largest size for which an exhaustive sample of the configuration space was computationally feasible in our implementation. Metrics are generally calculated as averages of 30 independent instances for each $K$ value.

\paragraph{Basins of attraction.}
\index{Basin of Attraction}

We start by analyzing the structure of the basins of attraction, namely, their size, shape and fitness of the corresponding local optima. These features are independent of the network edge definition. They are, however, relevant as the time complexity of local search heuristics is known to be linked to the size and spreading of attraction basins \cite{garnier01}.

\begin{figure}[htb!]
\begin{center}
\includegraphics[width=0.49\textwidth]{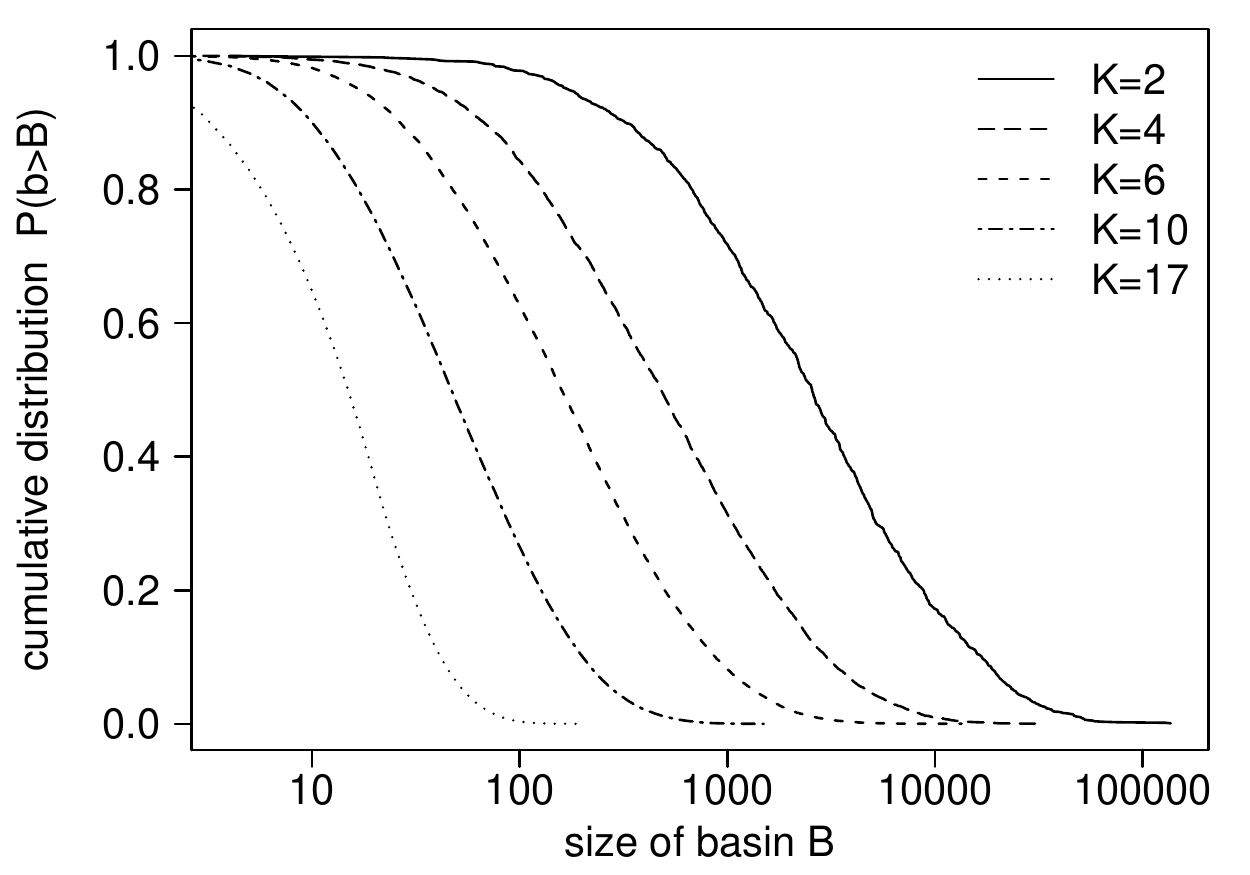}
\hspace{0.005\textwidth}
\includegraphics[width=0.49\textwidth]{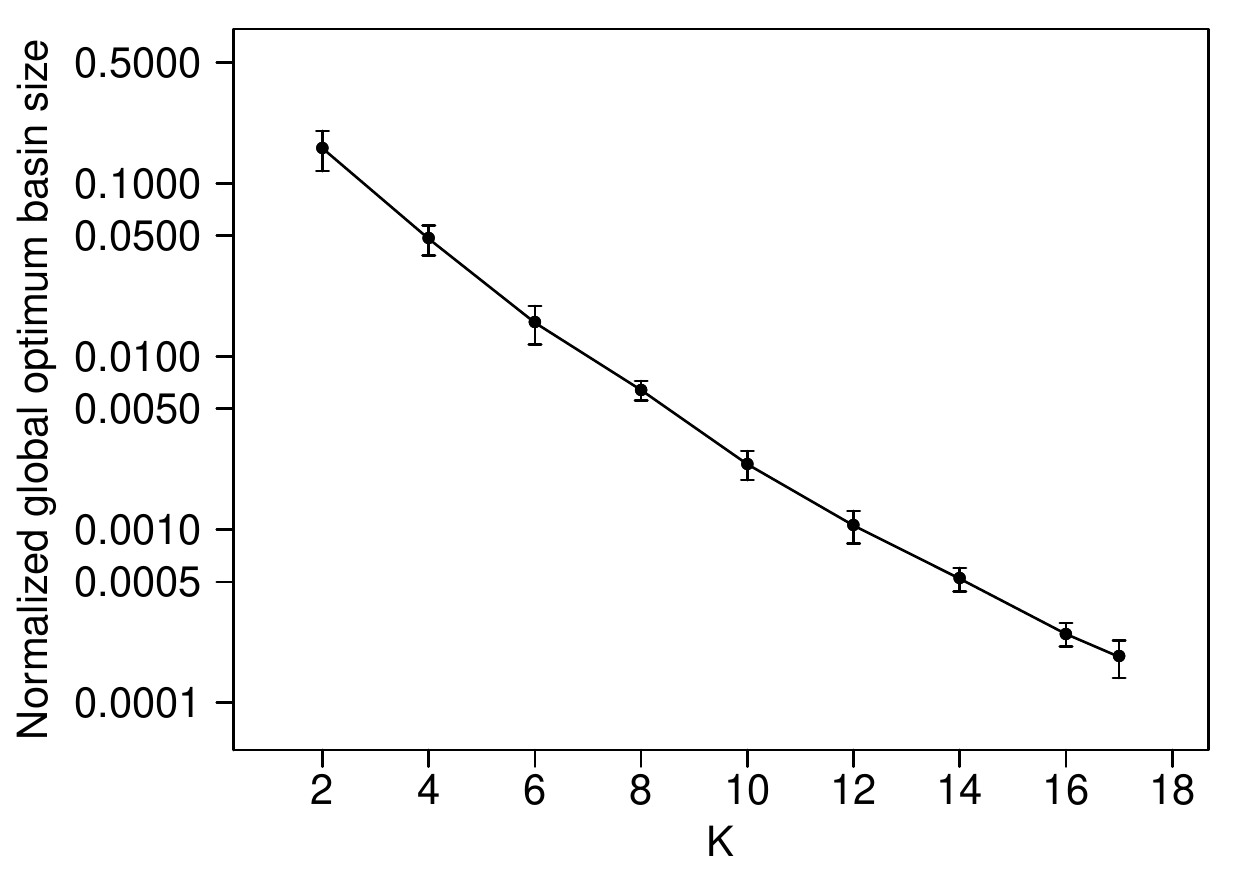}
\caption{Size of the basins of attraction for $NK$ Landscapes. {\bf Left}: cumulative distribution of basin sizes for landscapes with $N$ = 18, and selected values of $K$. {\bf Right}: average normalized size of the basin of the global optimum. Averages (points) and standard deviation (bars) refer to $30$ instances for each $K$ value.\label{fig:bas-size}}
\end{center}
\end{figure}

The distribution of basin sizes for given $N$ and $K$ values is not uniform; instead it follows a right-skewed distribution with a faster-than-exponential decay (see Figure~\ref{fig:bas-size}, left, with semi-logarithmic scale). With increasing ruggedness ($K$ values), the distribution shifts to the left and decays faster. This suggest that as the landscape ruggedness increases, the basin sizes decrease. In particular, with increasing ruggedness, the decrease of the relative size of global optimum basin is approximately exponential (Fig.~\ref{fig:bas-size}, Right).

\begin{figure}[htb!]
\begin{center}
\includegraphics[width=0.49\textwidth]{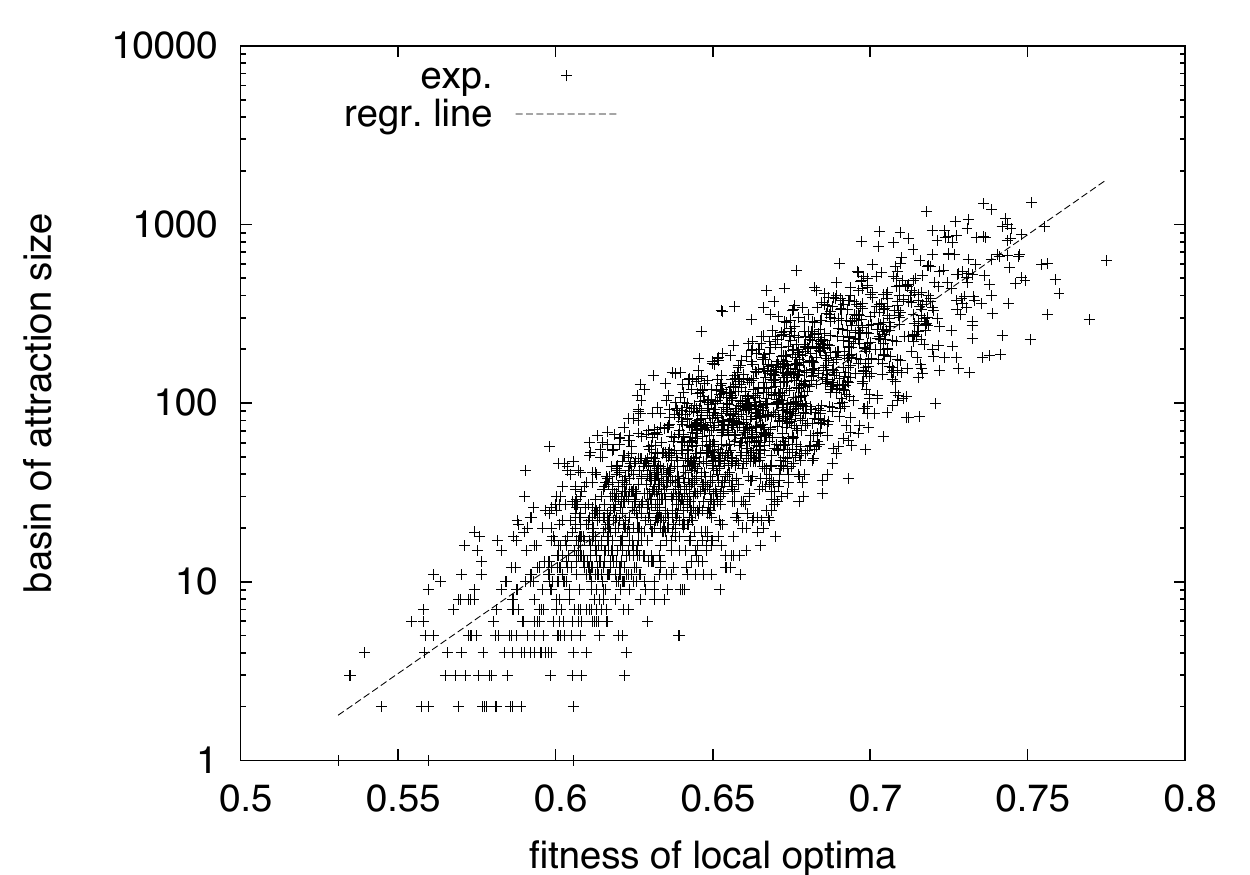}
\hspace{0.005\textwidth}
\includegraphics[width=0.49\textwidth]{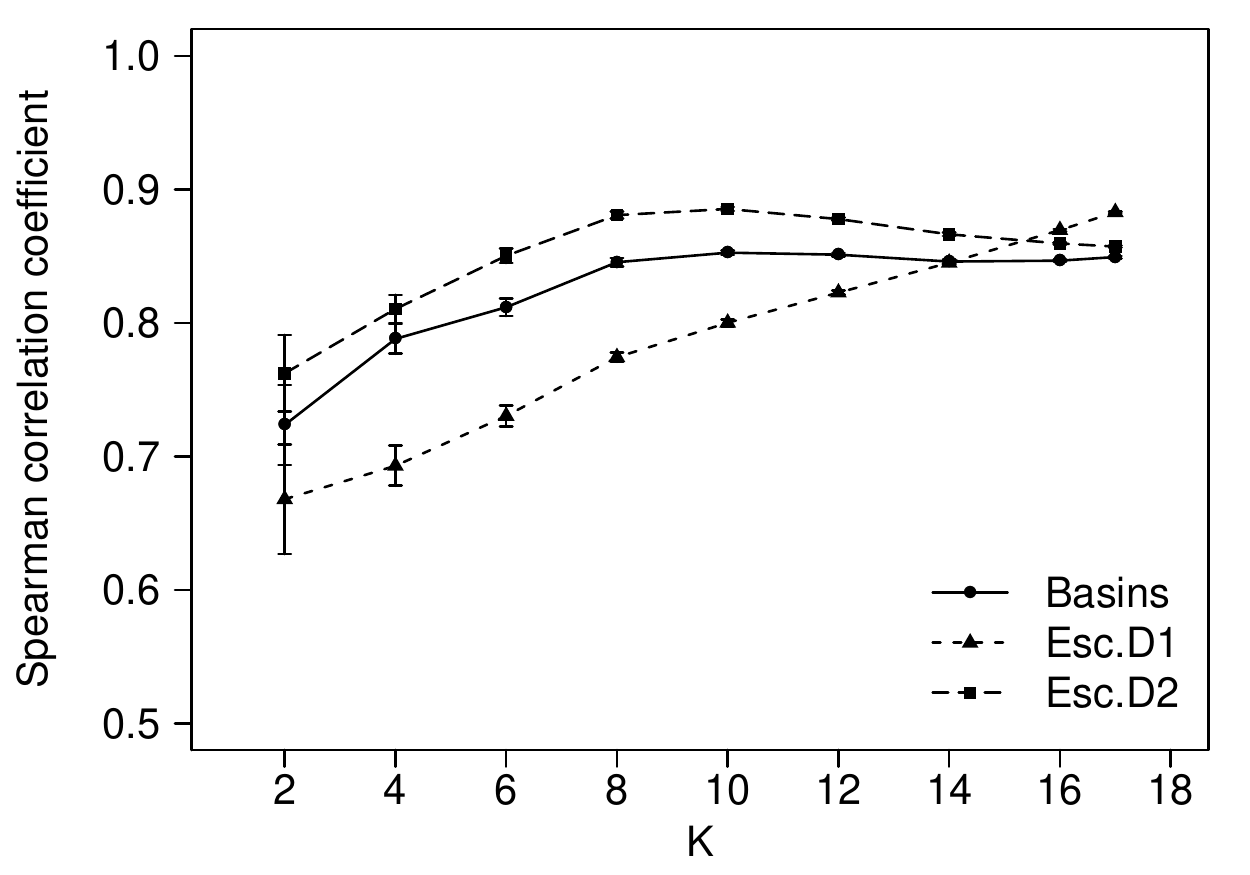}
\caption{Fitness correlations for $NK$ landscapes. {\bf Left}: Basin size and the fitness of its corresponding optima for a representative instance with $N=18, K=8$. {\bf Right}: fitness of an optimum and its strength, i.e. the sum of the weights of its incoming transitions. Averages (points) and $0.95$ confidence intervals (bars) are estimated by a t-test over $30$ instances. \label{fig:bas-cor}}
\end{center}
\end{figure}

With respect to the fitness of local optima and the size of their basins, a strong positive correlation was observed.  Surprisingly, the average Spearman correlation coefficient is above $0.8$ for all $K$ values. Figure~\ref{fig:bas-cor} (left) provides an example for $N =18$, $K=8$. This is an encouraging feature suggesting that local optima with high fitness should be easier to locate by hill-climbing. $NK$ landscapes can thus be imagined as mountain ranges where wider mountain basins belong to higher peaks. But intuitions can be misleading, a striking finding is that these mountains are hollow; for all the observed instances, the average size of the basin interior is always less than $1\%$ of the size of the basin itself. In other words, most solutions sit on the basin frontier and neighboring basins are richly interconnected~\cite{pre08}.

\paragraph{General network features.}

\begin{table}[!ht]
\begin{center}
 \caption{General network features for $NK$ landscapes. $K$ = epistasis value of the corresponding $NK$ landscape ($N=18$); $N_v$ = number of vertices; $D_{edge}$ = density of edges ($N_e / (N_v)^2 \times 100 \%$); $L_{opt}$ = average shortest path to reach the global optimum ($d_{ij}=1/w_{ij}$). Values are averages over 30 random instances, standard deviations are shown as subscripts. \label{tab:stats}}
 \begin{tabular}{ccccccccccc}
\toprule
\multirow{2}{*}{$K$}&\multicolumn{1}{c}{$N_v$}&\multicolumn{3}{c}{$D_{edge}~(\%)$}&\multicolumn{3}{c}{$L_{opt}$}\\
\cmidrule(lr){2-2}\cmidrule(lr){3-5}\cmidrule(lr){6-8}
&\multicolumn{1}{c}{all} &Basin-trans. &Esc.D1 &Esc.D2 &Basin-trans. &Esc.D1 &Esc.D2\\
\midrule
$2$&~~~$43.0_{27.7}$&$74.182_{13.128}$&$8.298_{4.716}$&$22.750_{9.301}$&~$21.2_{8.0}$&$16.8_{4.7}$&~$33.5_{14.1}$\\
$4$&~~$220.6_{39.1}$&$54.061_{4.413}$&$1.463_{0.231}$&~$7.066_{0.810}$&~$41.7_{10.5}$&$19.2_{5.1}$&~$53.7_{12.4}$\\
$6$&~~$748.4_{70.2}$&$26.343_{1.963}$&$0.469_{0.047}$&~$3.466_{0.279}$&~$80.0_{19.1}$&$22.2_{3.9}$&~$66.7_{12.9}$\\
$8$&~$1668.8_{73.5}$&$12.709_{0.512}$&$0.228_{0.009}$&~$2.201_{0.066}$&$110.1_{13.8}$&$24.0_{4.9}$&~$76.6_{9.1}$\\
$10$&~$3147.6_{109.9}$&~$6.269_{0.244}$&$0.132_{0.004}$&~$1.531_{0.036}$&$152.8_{19.3}$&$27.3_{5.0}$&~$90.7_{8.4}$\\
$12$&~$5270.3_{103.9}$&~$3.240_{0.079}$&$0.088_{0.001}$&~$1.115_{0.015}$&$185.1_{23.8}$&$30.3_{6.7}$&$108.3_{12.3}$\\
$14$&~$8099.6_{121.1}$&~$1.774_{0.035}$&$0.064_{0.001}$&~$0.838_{0.009}$&$200.2_{16.0}$&$38.9_{9.6}$&$124.7_{8.6}$\\
$16$&$11688.1_{101.3}$&~$1.030_{0.013}$&$0.051_{0.000}$&~$0.647_{0.004}$&$211.8_{15.0}$&$47.9_{11.4}$&$146.2_{11.2}$\\
$17$&$13801.0_{74.1}$&~$0.801_{0.007}$&$0.047_{0.000}$&~$0.574_{0.002}$&$214.3_{17.5}$&$55.7_{12.5}$&$155.9_{12.2}$\\
 \bottomrule
 \end{tabular} 
 \vspace{-0.3cm}
\end{center}
\end{table}

Table~\ref{tab:stats} reports some general features for the two network models: basin-transition edges and escape edges (with distances ($D = \{1, 2\}$); specifically, the number of nodes (which is independent of the edges model), the relative number or density of edges, and the average path length to the global optimum, where the distance between two nodes $i$ and $j$ is given by $1/w_{ij}$.

The number of local optima (Table~\ref{tab:stats}, $2^{nd}$ column) rapidly increases with the value of $K$ ($1^{st}$ column). Escape edges produce much less dense networks ($3^{rd}$, $4^{th}$, and $5^{th}$ columns), which confirms the visual inspection of Figures~\ref{fig:lonb} and \ref{fig:lone}. For all the models, the density of edges decreases, whereas the path length to the global optimum ($6^{th}$,$7^{th}$, and $8^{th}$ columns) increases with increasing values of $K$. Since a low density of edges and a long path length to the optimum would hinder heuristic search, these observations confirm that the network metrics capture the search difficulty associated with increasing landscape ruggedness. These findings also suggest that the two models of edges are consistent, which is encouraging as calculating the escape-edges is less computationally expensive.

\begin{figure}[htb!]
\begin{center}
\includegraphics[width=0.49\textwidth]{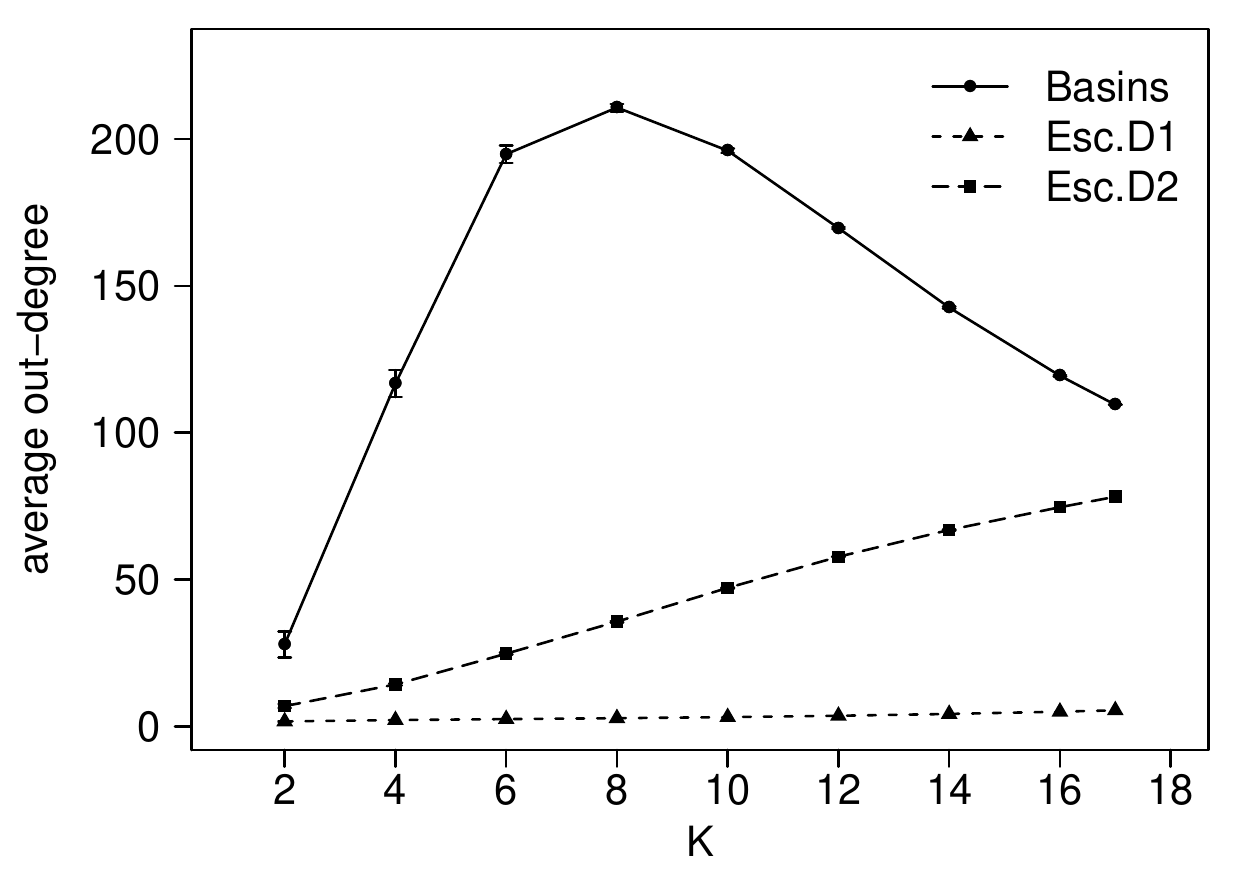}
\hspace{0.005\textwidth}
\includegraphics[width=0.49\textwidth]{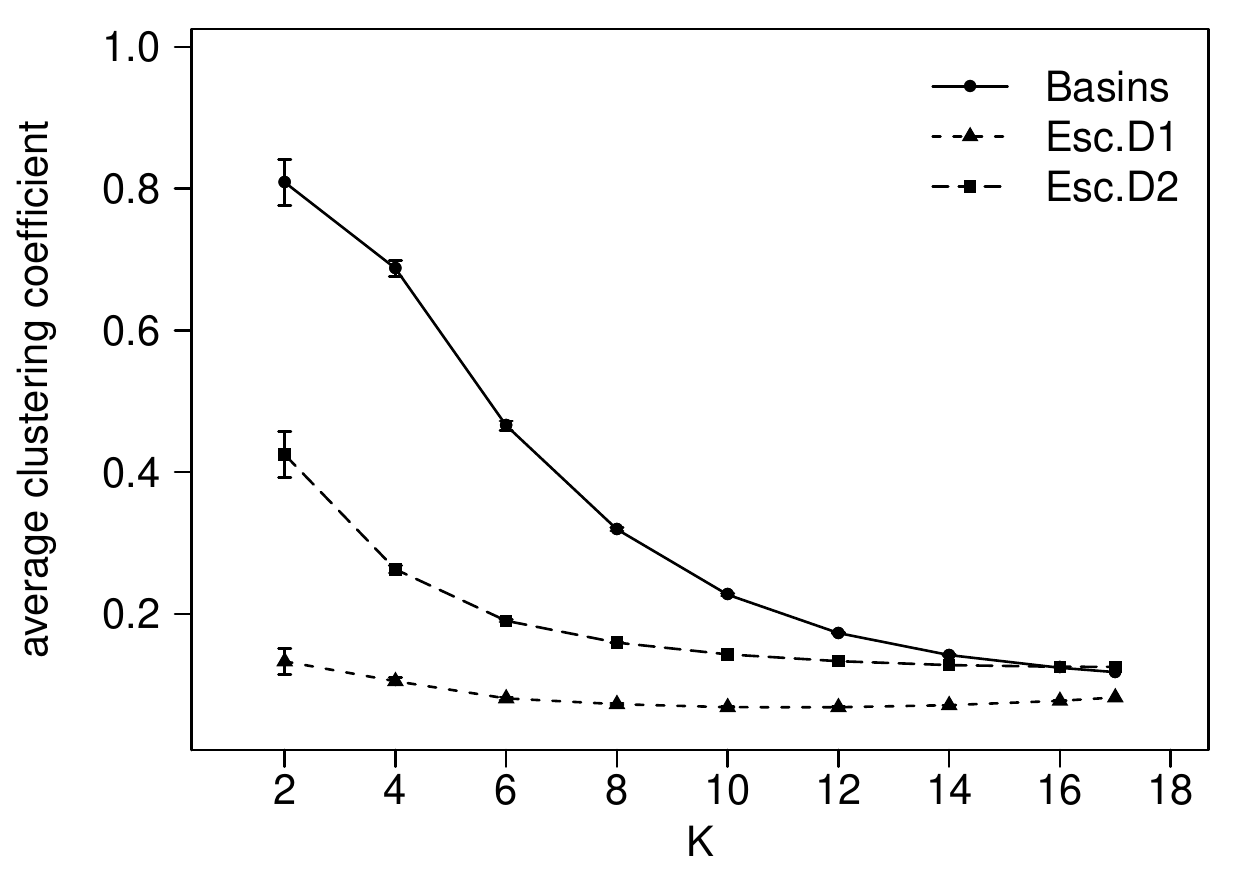}
\caption{Local connectivity for $NK$ landscapes. {\bf Left}: Average out-degree. {\bf Right}: Average clustering coefficient. Averages (points) and $0.95$ confidence intervals (bars) are estimated by a t-test over $30$ random instances with $N=18$.\label{fig:zout-cc}}
\end{center}
\end{figure}

A study of the network's local connectivity shows differences between the two edge models. As Fig.~\ref{fig:zout-cc} (left) shows, the basin-transition edges produce networks with higher out degree (i.e. number of transitions leaving a node). Clustering coefficients are also higher in this case  (they are indeed higher than those of a random graph), which is probably due to the higher density of basin-transition edges (Fig.~\ref{fig:zout-cc}, right).  There is, however, a common decreasing trend for all models in this metric with increasing $K$, as seen in Fig.~\ref{fig:zout-cc}, right. The varying difference between the two models might lie in the size of the basins of attraction. For low $K$ values, large basins produce high edge density and thus high clustering coefficients in the basin-transition model; whereas for large values of $K$, basins are so small that the two models show a similar structure. The escape-edges reproduce the basin topology.

\paragraph{Transitions among local optima.}

\begin{figure}[htb!]
\begin{center}
\includegraphics[width=0.49\textwidth]{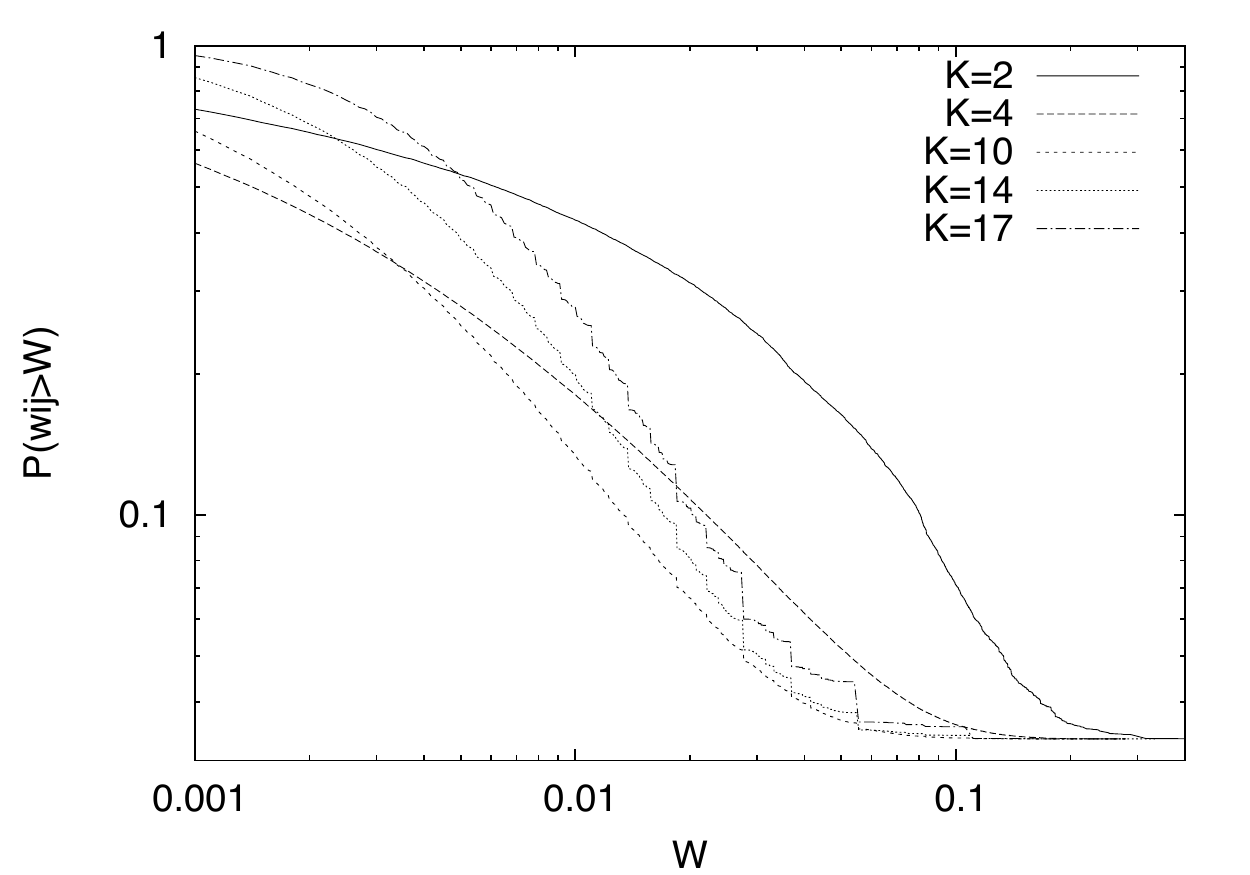}
\hspace{0.005\textwidth}
\includegraphics[width=0.49\textwidth]{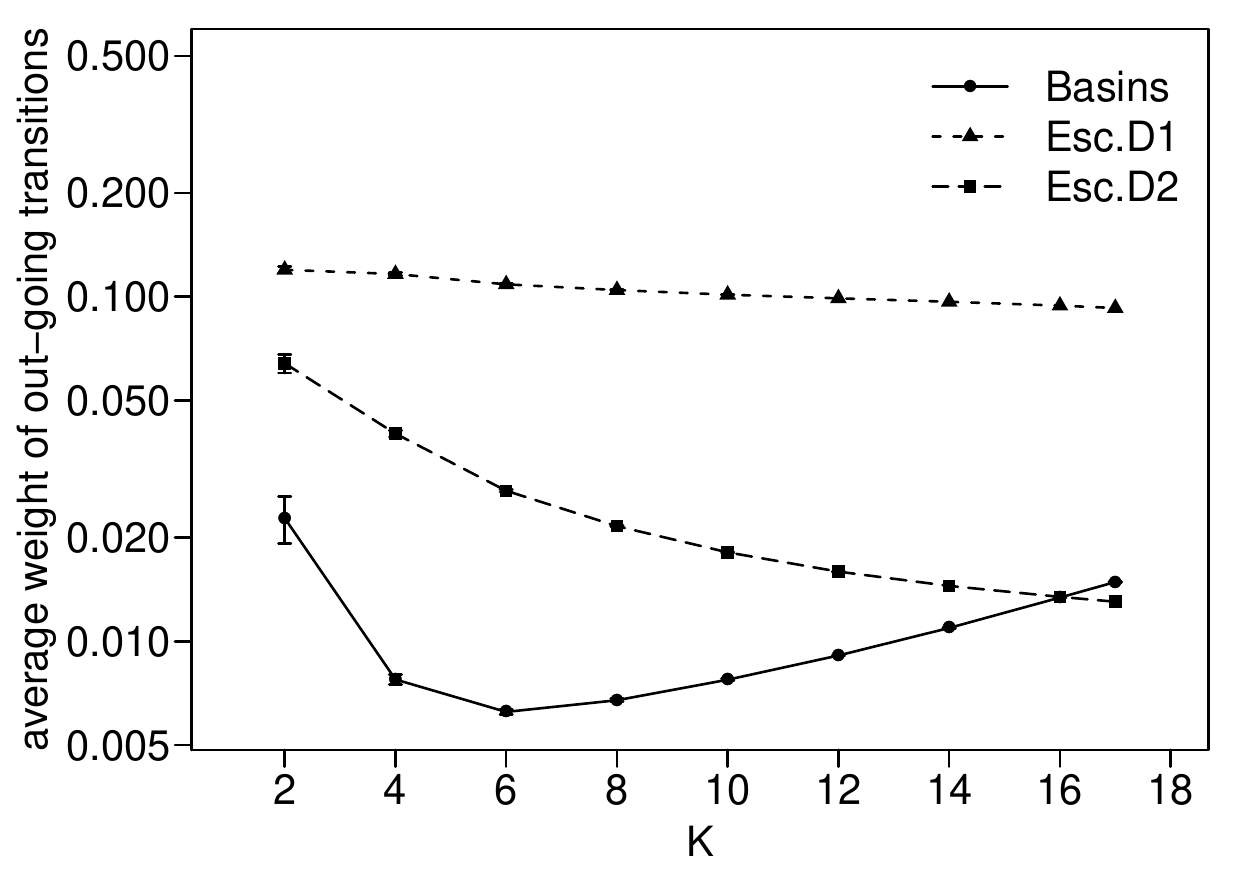}
\caption{Network transitions for $NK$ landscapes. {\bf Left}: Cumulative probability distribution of the network weights $w_{ij,j\ne i}$ with basin-transition edges, $N=18$, and selected $K$ values. {\bf Right}: Average out-going $w_{ij}$ values for all edge models and $K$ values. Averages (points) and $0.95$ confidence intervals (bars) are estimated by a t-test over $30$ landscapes with $N=18$.\label{fig:wij}}
\end{center}
\end{figure}

Edges weights $w_{ij}$ can be interpreted as the expected number of moves it takes to go from basin $b_i$ to basin $b_j$ (or from local optimum $i$ to basin $b_j$ in the escape-edge model). For both edge models, the weights of self-loops ($w_{ii}$) are an order of magnitude higher than $w_{ij,j \ne i}$. Therefore, it is more probable for a random move to remain in the same basin than to escape from it. Self-loop probabilities are then correlated with basin sizes, and display a similar exponential decrease with the landscape ruggedness $K$. We analyze, therefore, in more detail the weights $w_{ij,j \ne i}$. Figure~\ref{fig:wij} (left) shows the cumulative distribution of basin-transition weights for  $w_{ij,j \ne i}$ for selected values of $K$. The curves illustrate that low $K$ values have longer tails, whereas mid and high $K$ values produce a faster decay. Figure~\ref{fig:wij} (right), shows the average weight out-going transition for all edge models and $K$ values. For the escape-edges model, the out-going weights decrease smoothly, with a slower decrease for $D=1$. The trend is different for basin-transition edges where the out-going weights decrease with increasing ruggedness but only up to $K=6$, and then they increase in value.

\begin{figure}[htb!]
\begin{center}
\includegraphics[width=0.49\textwidth]{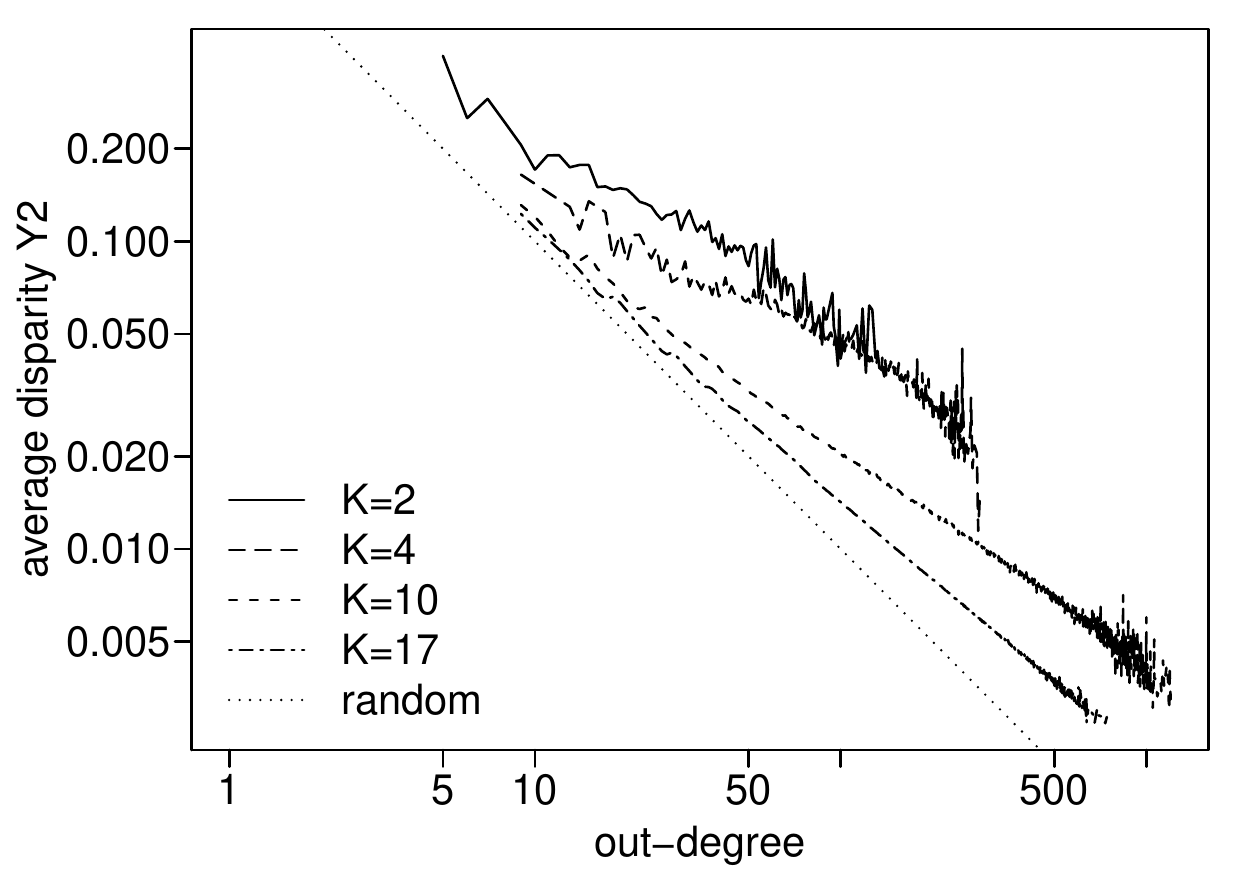}
\hspace{0.005\textwidth}
\includegraphics[width=0.49\textwidth]{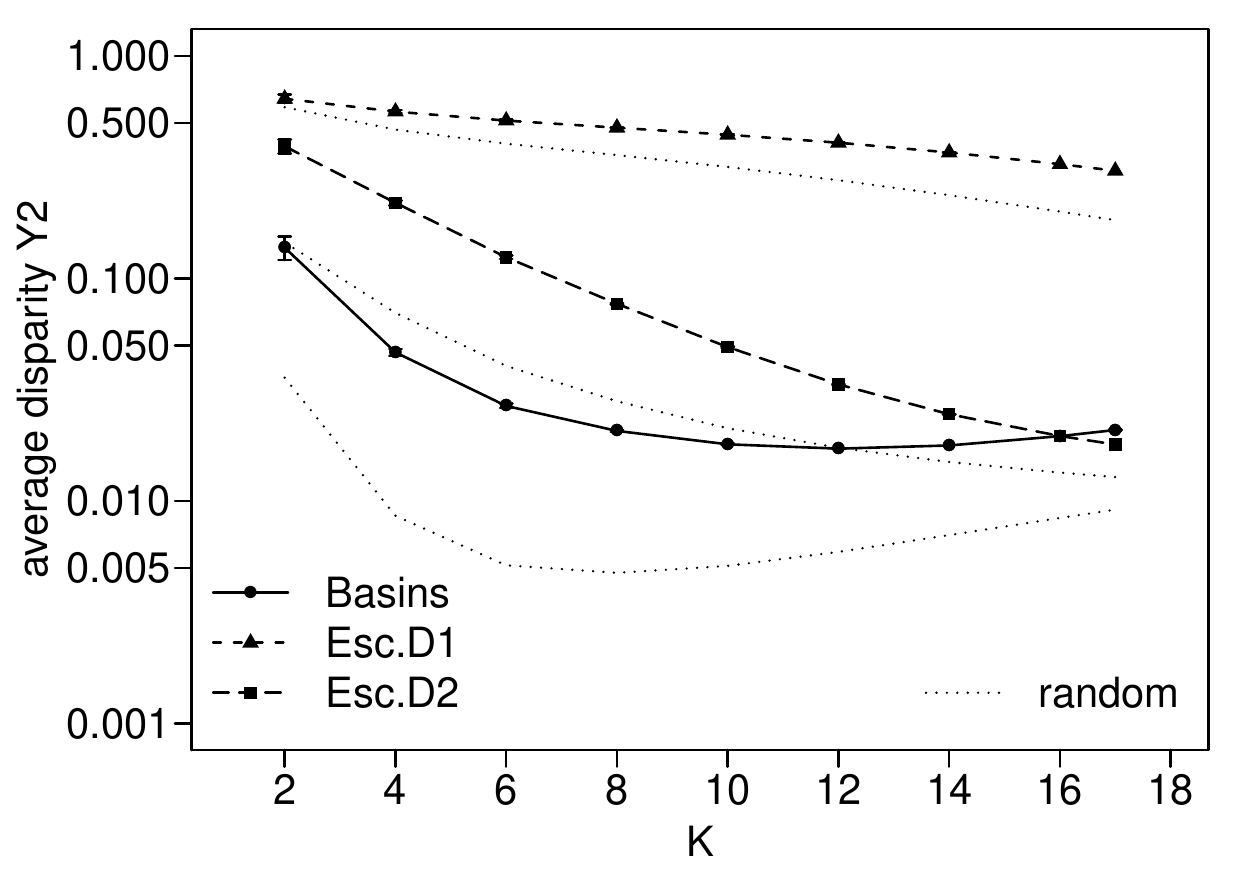}
\caption{Weight disparity for outgoing edges in $NK$ landscapes. {\bf Left}: relationship between disparity and out-degree for the basin-transition edges, and selected values of $K$. {\bf Right}: relationship between disparity and landscape ruggedness ($K$) for the two network models. Averages and confidence intervals are estimated on the $30$ analyzed instances. Dotted lines (labeled as random) present the limit case $Y_2=1/k$, where $k$ is the node out-going degree.\label{fig:y2}}
\end{center}
\end{figure}

A relevant question is whether there are preferential directions when leaving a particular node in the network. Specifically, whether for a given optimum $i$, all the outgoing weights $w_{ij,j \ne i}$, are equivalent. This can be revealed by the disparity $Y_2$ metric (discussed in section \ref{WN}), which gauges the heterogeneity of the contributions of the edges of node $i$ to its total strength $s_i$. If a dominant weight does not exist, the value  $Y_2 \approx 1/k$, were $k$ is the node out-degree. Figure~\ref{fig:y2} (left) illustrates the relationship between disparity and out-degree for basin-transition edges and selected $K$ values. The figure also shows the limit case $Y_2 \approx 1/k$ (labeled as random). For calculating this plot, the nodes' disparity values $Y_2(i)$ were grouped and averaged by node out-degree. The curves suggest that there are preferential directions for low values of $K$. However,  with increasing $K$, the transition probabilities to leave a given basin appear to become more uniform (i.e closer to the limit case $Y_2=1/k$).  Figure~\ref{fig:y2} (right) shows the disparity metric for all models and $K$ values. In all cases, disparity values are higher than those expected in the limit case ($Y_2 \approx 1/k$, labeled as random), indicating that preferential transitions are present. For the basin-transition edges and the escape edges wit $D=2$, disparity values get closer to the limit case of large $K$ values.

\paragraph{Search difficulty and network metrics.}
\index{Search Difficulty}

While the previous sections described relevant network features, this section explores correlations between these features and the performance of local search heuristics running on the underlying combinatorial optimization problem. The ultimate goal is to have predictive models of the performance of specific search heuristics when solving a given problem instance, and thus select a method according to this predication.

Daolio et al ~\cite{daolio2012local}, conducted a first study using {\em iterated local search} and the $NK$ family of landscapes (with escape edges,  $D=2$). Iterated local search is a relatively simple but powerful single point heuristic. It alternates between a perturbation stage and an improvement stage. This search principle has been rediscovered multiple times, within different research communities and with different names \cite{Battiti2009}. The term {\em iterated local search} (ILS) was proposed in \cite{lourenco:2002}.  Algorithm~\ref{alg:ILS} outlines the procedure.

\begin{algorithm}[!h]
\caption{Iterated Local Search} \label{alg:ILS}
$s_0 \leftarrow \text{GenerateInitialSolution}$\;
$s^* \leftarrow \text{LocalSearch}(s_0)$\;
\Repeat{termination condition met}{
	$s' \leftarrow \text{Perturbation}(s^*)$\;
	$s'^{*} \leftarrow \text{LocalSearch}(s^{'})$\;
	$s^* \leftarrow \text{AcceptanceCriterion}(s^*,s'^{*})$\;
}
\end{algorithm}

In our implementation,  the {\em LocalSearch} stage corresponds to  the best-improvement hill-climber described in section \ref{lon} (Algorithm~\ref{algoHC}), which stops when reaching a local optimum, and uses the single bit-flip move operator. The {\em Perturbation} stage considers a stronger operator,i.e. 2-bit-flip mutation. A simple greedy acceptance is used (i.e. only improvement moves are accepted). The search terminates at the global optimum, which for benchmark problems is known \textit{a priori}, or when reaching a pre-set limit of fitness evaluations $FE_{max}$.

As the performance criterion, we selected the expected number of function evaluations to reach the global optimum (\emph{success}) after independent restarts. This measure accounts for both the success rate ($p_s \in (0,1]$) and the convergence speed~\cite{auger2005performance}. The function evaluations limit was set to $1/5$ of the size of the search space, i.e. $FE_{max} \simeq 5.2 \cdot 10^{4}$, for binary strings of length $N=18$. The success rate $p_s$ and running time of successful runs $T_s$ were estimated on $500$ random restarts per instance.

\begin{figure}[htb!]
\begin{center}
\includegraphics[width=0.49\textwidth]{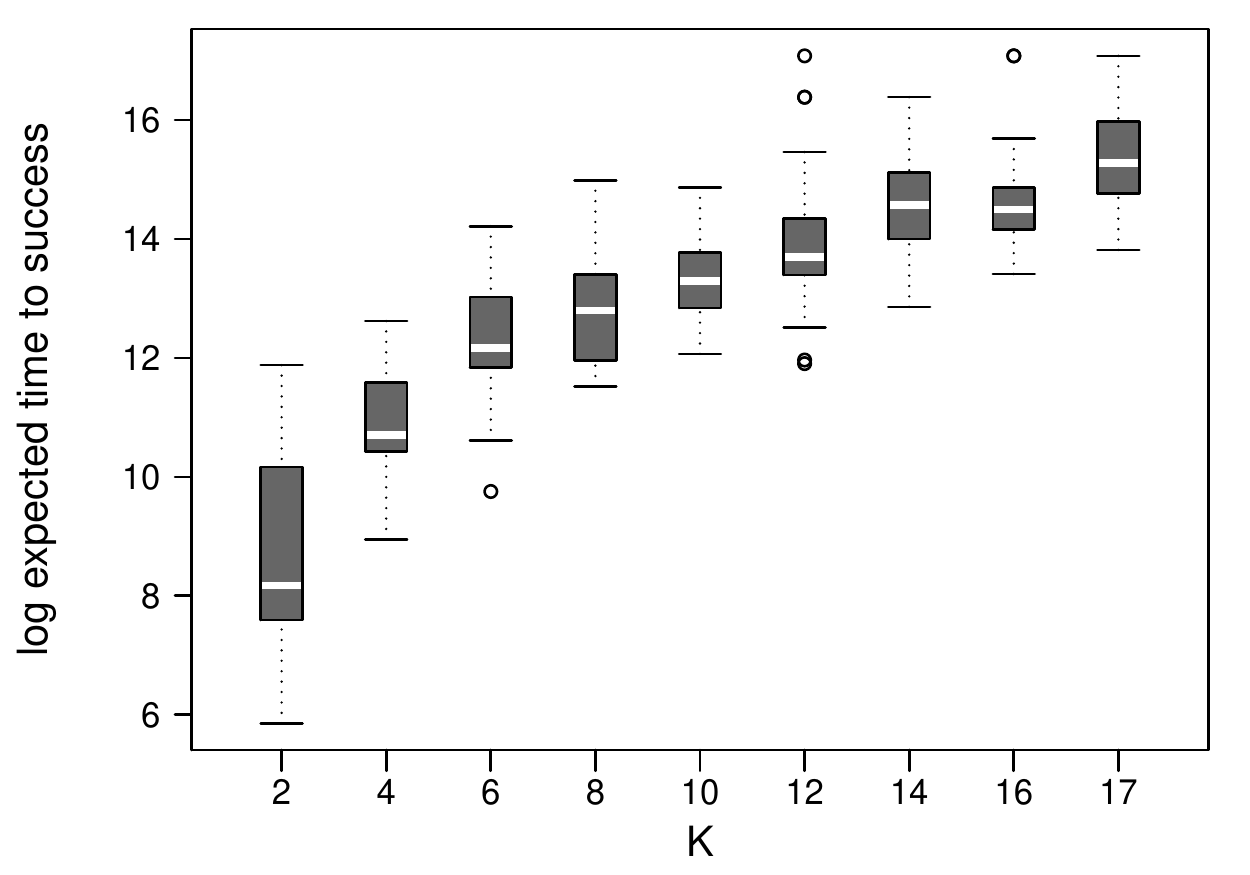}
\hspace{0.005\textwidth}
\includegraphics[width=0.49\textwidth]{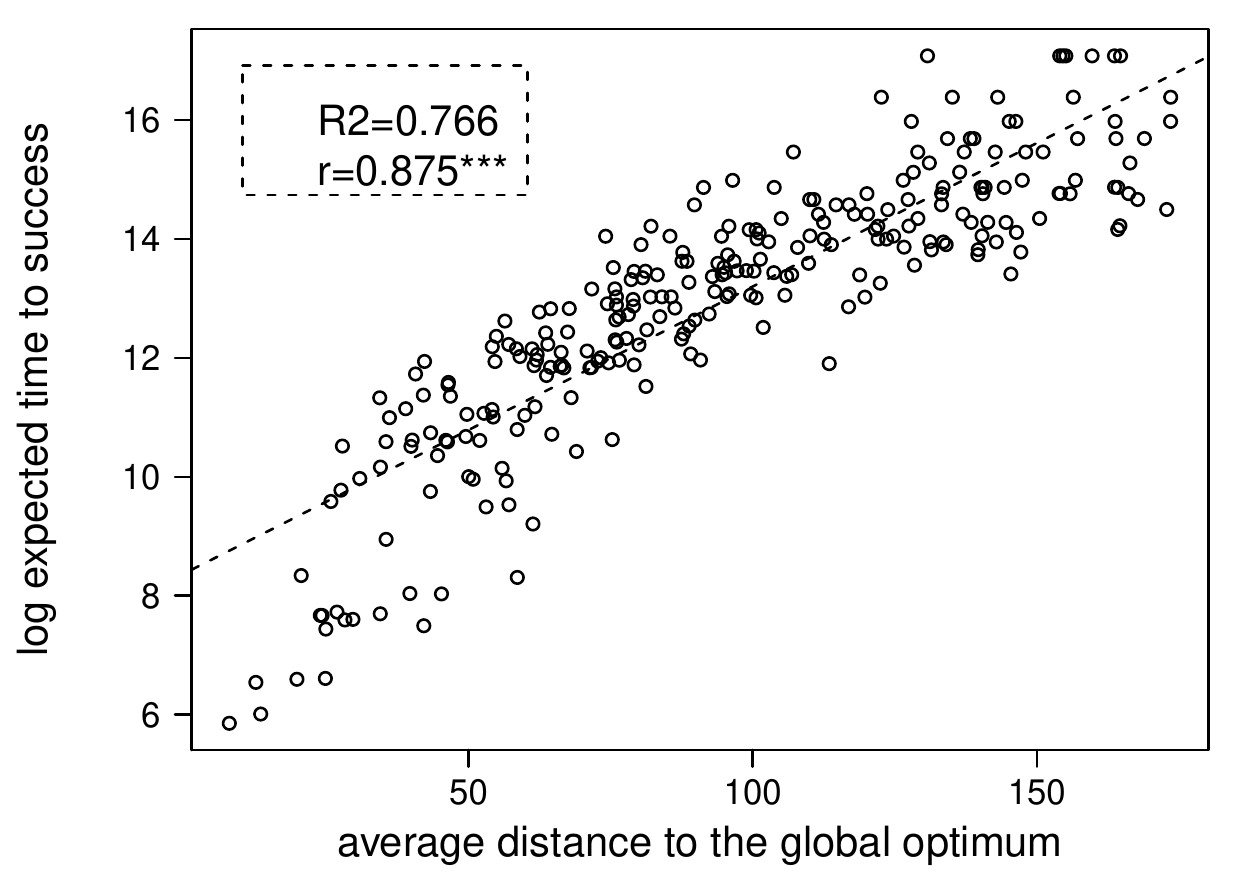}
\caption{Performance of ILS on $NK$ landscapes. {\bf Left}: distribution of the expected running times to success for different $K$ values. {\bf Right}: correlation  between the expected running times and the average shortest path to the global optimum. The regression line is dashed. The legend gives the ratio of variance explained by the regression, $R2$, and the Pearson correlation coefficient, $r$, with the asterisks indicating its significance level. \label{fig:ILSperf}}
\end{center}
\end{figure}

Figure \ref{fig:ILSperf} (left), shows the distribution of the expected iterated local search (ILS, Algorithm~\ref{alg:ILS}) running times to success with respect to $K$. As expected, the running times increase steadily with increasing $K$. As an example of the correlations arising between LON features and the performance of ILS, the right plot in Figure.~\ref{fig:ILSperf}, illustrates the relationship between the running time  and the shortest path length to the global optimum. A strong positive correlation is observed, suggesting that LON features are able to capture search difficulty in combinatorial landscapes. Other network metrics also revealed positive correlations with search performance, namely, the average out-degree, the average disparity, and the degree of assortativity ~\cite{daolio2012local}. A multiple regression analysis was also conducted using the most significant network metrics. The model obtained was able to predict about $85\%$ of the  variance observed in the expected running times

\subsection{The Quadratic Assignment Problem}
\label{ResultsQAP}
\index{Quadratic Assignment Problem}
\index{QAP}

This section summarises the main results for the QAP problem. In this case, only the basin-transition edges are considered. The study of escape-edges will be the subject of future work. An analysis of the LON communities structure is also presented. This was not done for the $NK$ landscape as our analysis revealed little cluster structure of local optima in these more random landscapes. Their search spaces seem isotropic from the point of view of basin inter-connectivity. An initial study correlating QAP local optima network metrics with heuristic search performance is reported elsewhere ~\cite{chicano2012}.

Two QAP instance classes were considered: real-like and uniform instances as described in section \ref{sec:qapdef}. For the general network analysis, 30 random uniform and 30 random real-like instances have been generated for each problem dimension in $\{5,...,10\}$, and metrics are given as averages of these 30 independent instances. To the specific purpose of community detection, 200 additional instances have been produced and analyzed with size $9$ for the random uniform class, and size $11$ for the real-like instances class. Problem size $11$ is the largest permitting an exhaustive sample of the configuration space. Only the basin-transition edges are studied. The escape-edges will be the subject of future work.

\paragraph{Basins of attraction.}
\index{Basin of Attraction}

\begin{figure}[htb!]
\begin{center}
\includegraphics[width=0.49\textwidth]{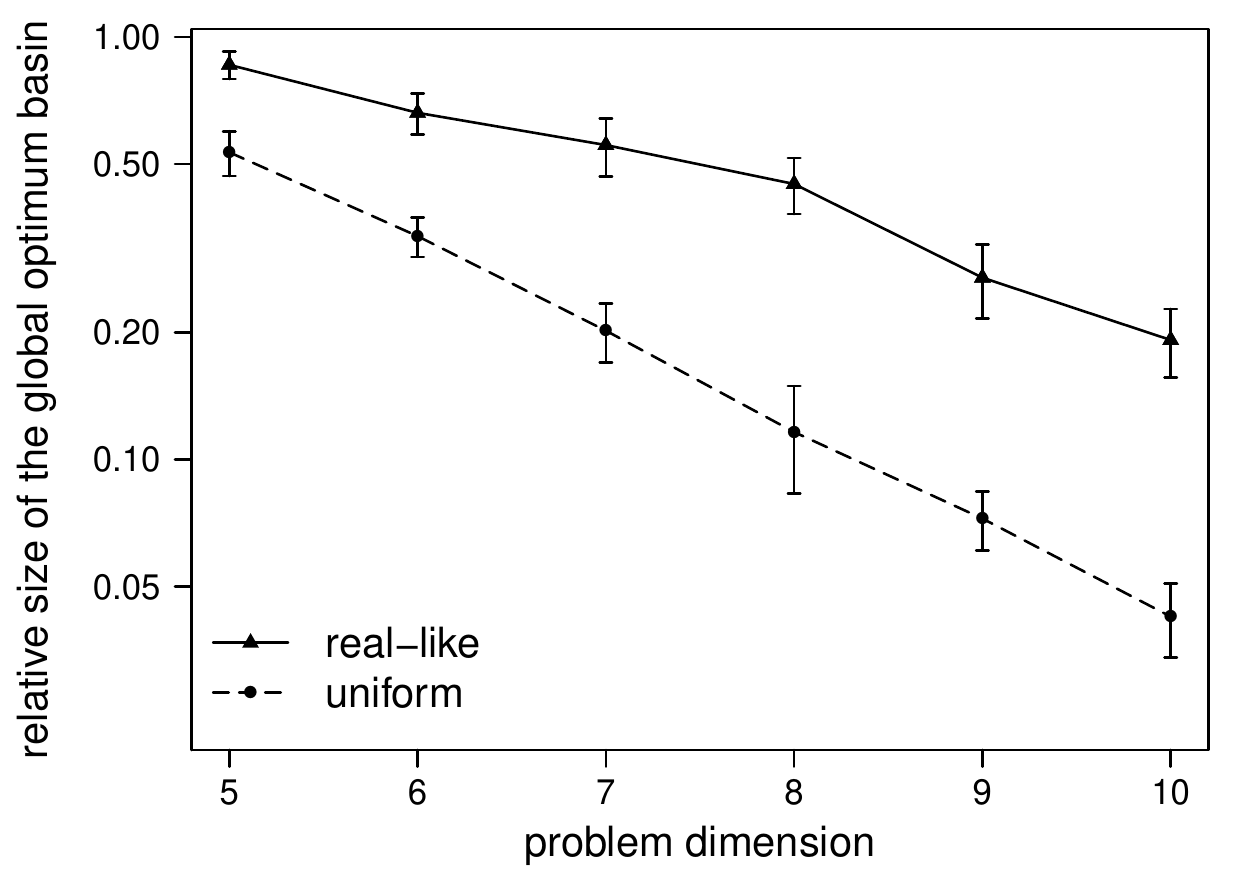}
\hspace{0.005\textwidth}
\includegraphics[width=0.49\textwidth]{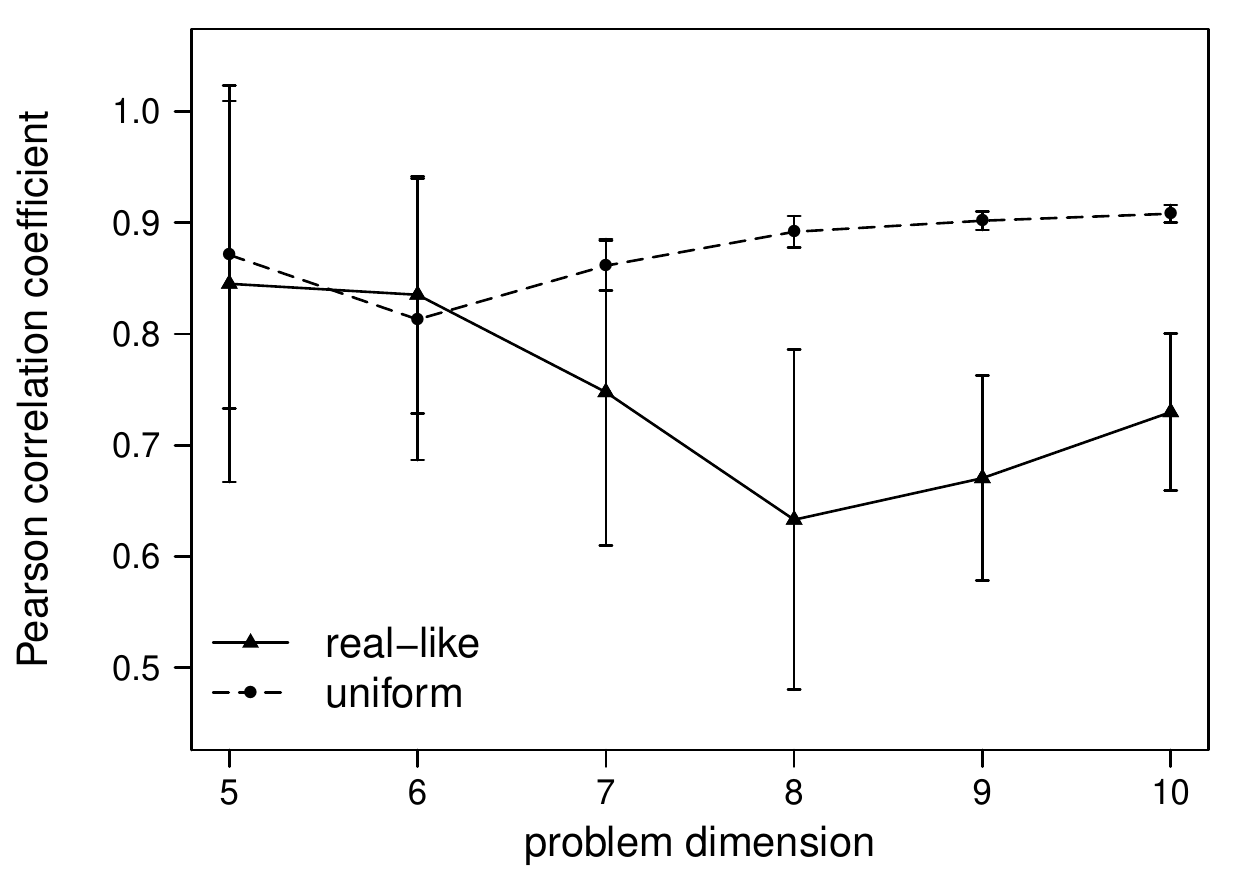}
\caption{Basins of attraction in QAP instances. {\bf Left}: Normalized size of the basin of the global optimum. {\bf Right}: Pearson correlation coefficient between the fitness value of an optimum and the logarithmic size of its basin. Averages (points) and $0.95$ confidence intervals (bars) are estimated with a t-test over $30$ instances for each combination of problem class and size.\label{fig:qap-bas}}
\end{center}
\end{figure}

Figure~\ref{fig:qap-bas} (left) shows the size of the global optimum basin of attraction (normalized by the whole search space size). This value decreases exponentially with the problem size for both instance classes. The real-like instances have larger global optimum basins, which can be explained by their smaller local optima networks (as discussed below). The relative size of the global optimum basin is related to the probability of finding the best solution with a local search algorithm from a random starting point. The exponential decrease confirms that the larger the problem, the smaller the probability for a local search algorithm to locate the global optimum. The separation between the curves in fig.~\ref{fig:qap-bas} (left) is consistent with recent empirical  results indicating that real-like instances are easier to solve than uniform instances for heuristic search algorithms such as simulated annealing and genetic algorithms \cite{chicano2012}.

Figure~\ref{fig:qap-bas} (right), shows the correlation between the fitness value of a local optimum and the size of its basin of attraction. As with the $NK$ landscape, there is a strong positive correlation between the fitness of a local optimum and the size of its basin. This is an encouraging feature suggesting that local optima with high fitness should  be easier to locate by stochastic local search. The correlation coefficients are generally higher for the uniform instances, which also show noticeably lower variance. More details about QAP basins of attraction and network features can be found in~\cite{cec10}.

\paragraph{General network features.}

Table~\ref{tab:qap-stats} reports relevant features for the two classes of QAP instances and problem sizes from 5 to 10; specifically, the  number of vertices ($N_v$),  the density of edges ($D_{edge} = N_e / (N_v)^2$), the weighted clustering coefficient ($C^w$) and the disparity in out-going transitions ($Y_2$). The number of local optima grows exponentially with the problem dimension for both instance classes.  For a given problem size, however, real-like instances produce much smaller networks, i.e. they have significantly fewer local optima.  The size difference between the two instance classes also grows almost exponentially with the problem dimension. This is again consistent with  the empirical studies indicating that real-like instances are easier to solve by common metaheuristics \cite{chicano2012}.

The QAP networks are notably dense, with density of edges close to one (Table~\ref{tab:qap-stats}, $D_{edge}$), much more dense than than those of the  $NK$ landscapes which operate on binary spaces. This is not surprising as the neighborhood size is larger for permutation search spaces (see Table \ref{tab:moves}). Local optima networks are almost complete graphs for QAP. Moreover, the average weighted clustering coefficient $C^w$ is higher than what would be expected from network density alone,  indicating that the interconnected triples are more likely formed by edges with larger weight. The studied QAP  instances show very high local connectivity. The clustering coefficient decreases with the problem dimension and is higher for real-like instances.

The last row in Table~\ref{tab:qap-stats} reports the disparity coefficient in out-going transitions for both classes with respect to the problem dimension. High diversity indicates preferential transitions. The decreasing trend reflects that, with increasing problem size, the out-going transition to neighbouring  optima tend to become equally probable. This trend is more evident for uniform instances whose LONs have higher cardinality.

\begin{table}[hptb]
\begin{center}
\caption{General network features for QAP instances. $N_v$ = number of vertices; $D_{edge}$ = density of edges ($N_e / (N_v)^2$); $C^w$ = weighted clustering coefficient; $Y_2$ = disparity in out-going transitions. Values are averages over $30$ instances with standard deviations in subscripts. \label{tab:qap-stats}}
\begin{tabular}{cccccccccc} \toprule
\multicolumn{1}{}{}&\multirow{2}{*}{class}&\multicolumn{6}{c}{size}\\
\cmidrule(lr){3-8}
\multicolumn{2}{}{}&\multicolumn{1}{c}{5}&\multicolumn{1}{c}{6}&\multicolumn{1}{c}{7}&\multicolumn{1}{c}{8}&\multicolumn{1}{c}{9}&\multicolumn{1}{c}{10}\\
\cmidrule{2-8}
\multirow{2}{*}{$N_v$}&real-like&~~$1.667_{0.802}$ &~~$2.767_{1.48}$&~~$3.900_{2.25}$&~~$6.133_{2.99}$ &~$12.567_{5.73}$ &~$25.700_{13.8}$\\
&uniform&~~$3.333_{1.27}$&~~$6.800_{2.37}$&~$19.100_{7.39}$&~$51.300_{20.53}$&$137.300_{54.84}$&$414.133_{177.5}$\\
\midrule
\multirow{2}{*}{$D_{edge}$}&real-like &$1.000_{0.000}$ &$0.993_{0.026}$&$0.994_{0.030}$&$0.999_{0.006}$&$0.992_{0.025}$&$0.988_{0.035}$\\
&uniform &$0.998_{0.007}$&$0.993_{0.019}$&$0.969_{0.030}$&$0.940_{0.036}$&$0.909_{0.035}$&$0.874_{0.053}$\\
\midrule
\multirow{2}{*}{$C^w$}&real-like &$1.000_{0.000}$ &$0.988_{0.032}$ &$0.995_{0.024}$ &$0.999_{0.005}$ &$0.995_{0.015}$ &$0.993_{0.020}$\\
&uniform &$0.998_{0.008}$ &$0.995_{0.014}$ &$0.982_{0.016}$ &$0.970_{0.017}$ &$0.961_{0.015}$ &$0.952_{0.020}$\\
\midrule
\multirow{2}{*}{$Y_2$}&real-like &$0.888_{0.193}$ &$0.739_{0.251}$ &$0.533_{0.270}$ &$0.369_{0.171}$ &$0.221_{0.118}$ &$0.143_{0.058}$\\
&uniform &$0.649_{0.271}$ &$0.286_{0.093}$ &$0.136_{0.061}$ &$0.074_{0.048}$ &$0.040_{0.013}$ &$0.023_{0.008}$\\
\bottomrule
\end{tabular}
\end{center}
\end{table}

\paragraph{Path lengths.}

\begin{figure}[htb!]
\begin{center}
\includegraphics[width=0.49\textwidth]{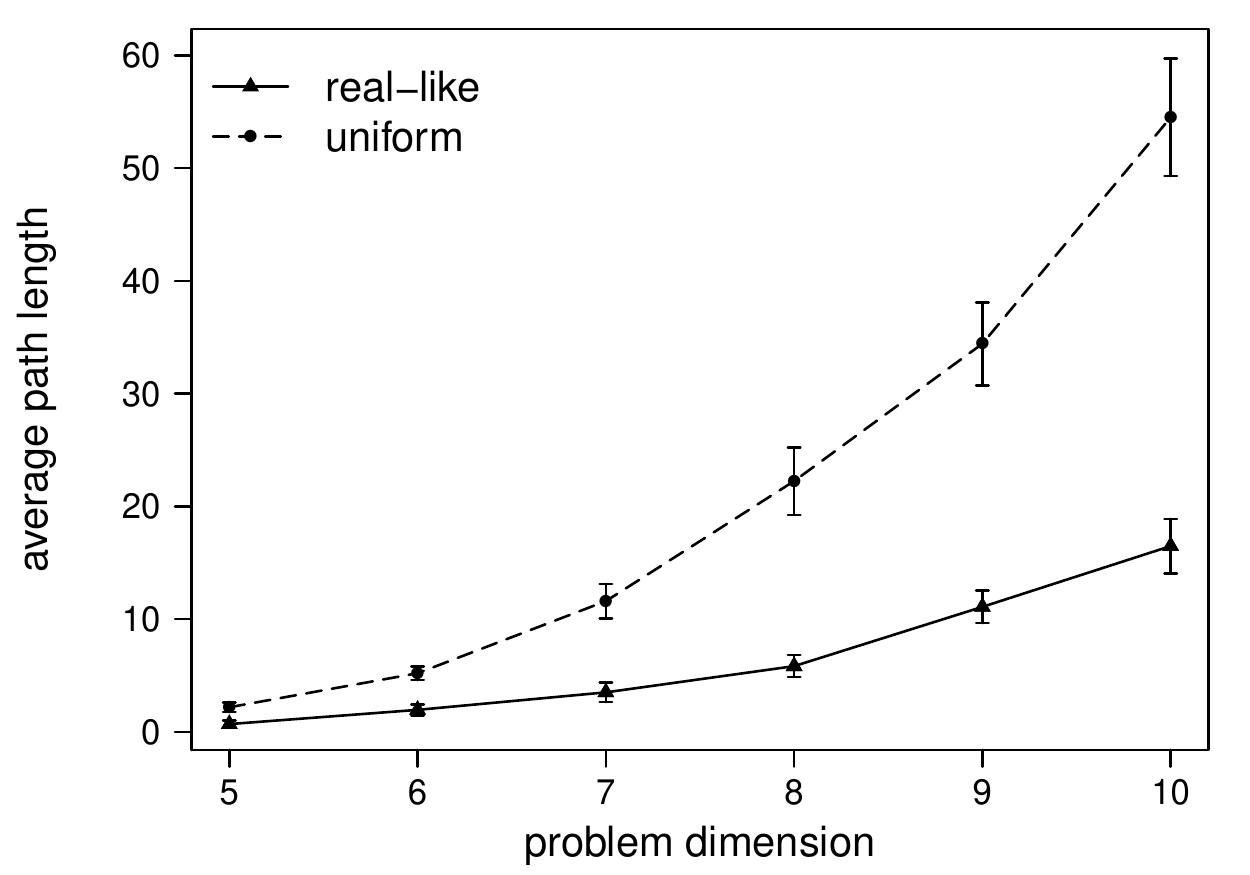}
\hspace{0.005\textwidth}
\includegraphics[width=0.49\textwidth]{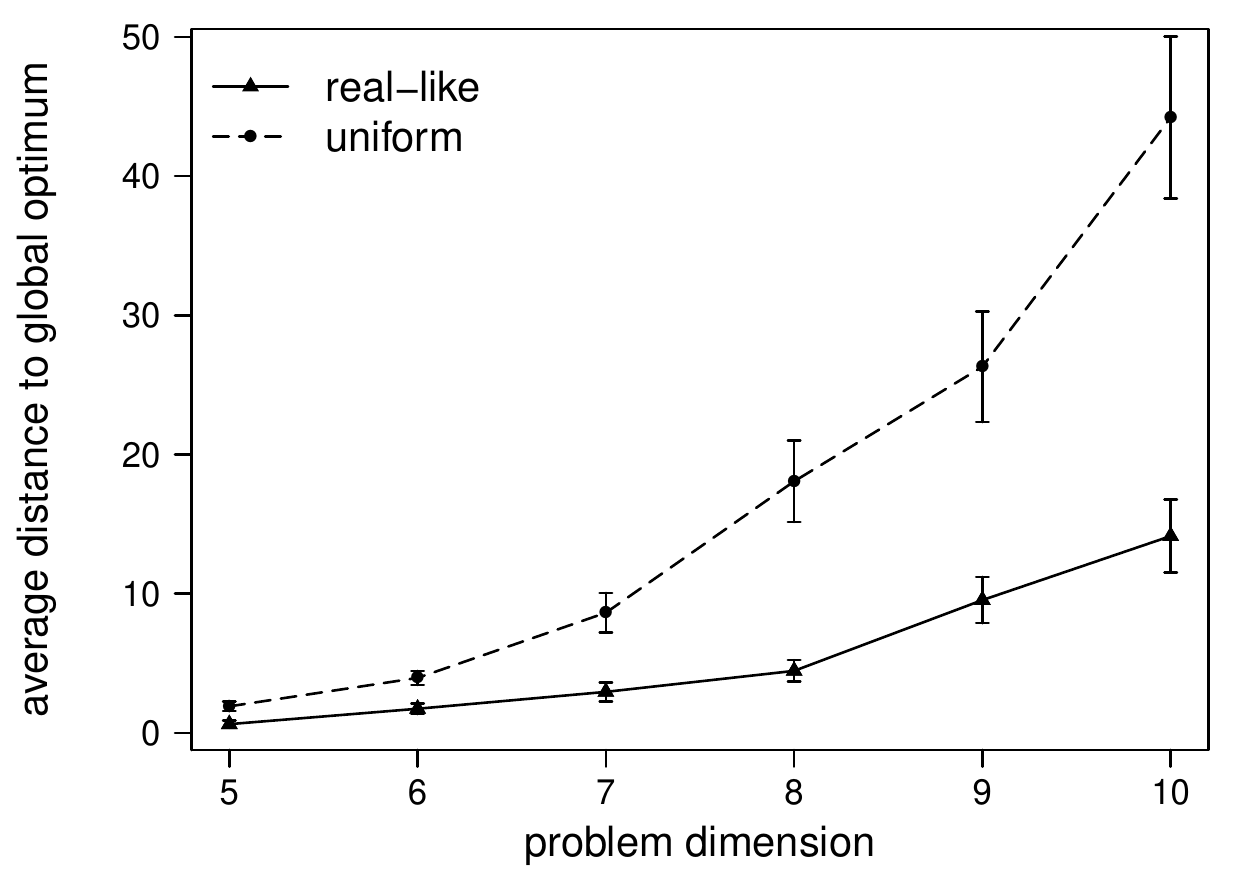}
\caption{Shortest paths in QAP instances. {\bf Left}: average path length. {\bf Right}: average length of the shortest paths to the global optimum. Averages (points) and $0.95$ confidence intervals (bars) are estimated with a t-test over $30$ instances for each combination of problem class and size.\label{fig:qap-paths}}
\end{center}
\end{figure}

Figure~\ref{fig:qap-paths}  (left )displays the average shortest path length between optima and the average shortest path length to the global optimum. Both metrics clearly increase with problem size. Values are noticeably higher for the uniform instances, which have a larger number of local optima than the real-like instances for the same problem dimension. The figures support that the search difficulty increases with the problem size and  the number of local optima.

\paragraph{Clustering of local optima.}

 The manner in which local optima are distributed in the configuration space is relevant for heuristic search. Several questions can be raised. Are they uniformly distributed, or do they cluster in some non-homogeneous way? If the latter, what is the relation between objective function values within and among different clusters and how easy is it to go from one to another? As discussed in section \ref{sec:community}, clusters or communities in networks are groups of nodes that are strongly connected between them and poorly connected with the rest of the graph. The topological distribution of local optima can be directly investigated by means of community detection on the local optima network. In ~\cite{daolio2011communities}, we conducted a community detection study on the two classes of QAP instances. Problems of size of  $11$ for the real-like class and $9$ for the uniform class were considered as LONs for these two cases have comparable sizes in terms of number of vertices.

\begin{figure}[h!]
\begin{center}
 \includegraphics[width=0.55\textwidth]{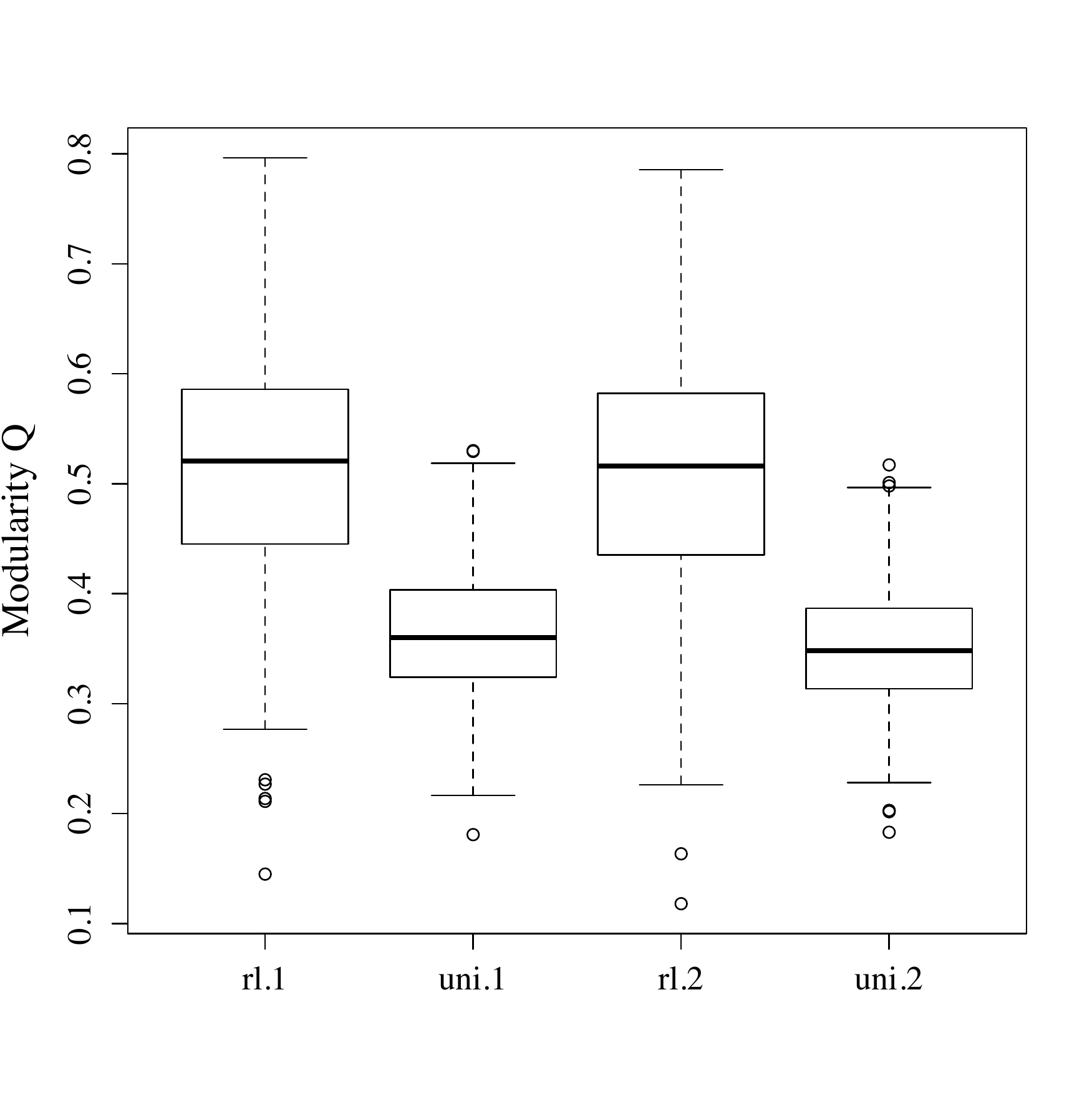}
  \vspace{-0.4cm}
 \caption{Network modularity for QAP instances. Boxplots of the modularity score $Q$ on the y-axis with respect to class problem (rl stands for real-like and  uni stands for random uniform) and community detection algorithm (1 stands for fast greedy modularity
  optimization and 2 stands for spin glass search algorithm).}
\label{fig:box-qap}
\end{center}
\end{figure}

Community detection is a difficult task, but today several good approximate algorithms are available~\cite{santo1}. In ~\cite{daolio2011communities},   we used two of them: (i) a method based on greedy modularity optimization, and (ii) a spin glass ground state-based algorithm in order to double check the community partition results. Figure~\ref{fig:box-qap} shows the modularity score ($Q$) distribution for each algorithm/instance-class.  The higher the $Q$ score of a partition, the crisper the community structure~\cite{santo1}.  The plot indicates that the two instance classes are well separated in terms of $Q$, regardless of the algorithm used.

\begin{figure}[htb!]
\begin{center}
\includegraphics[width=0.49\textwidth]{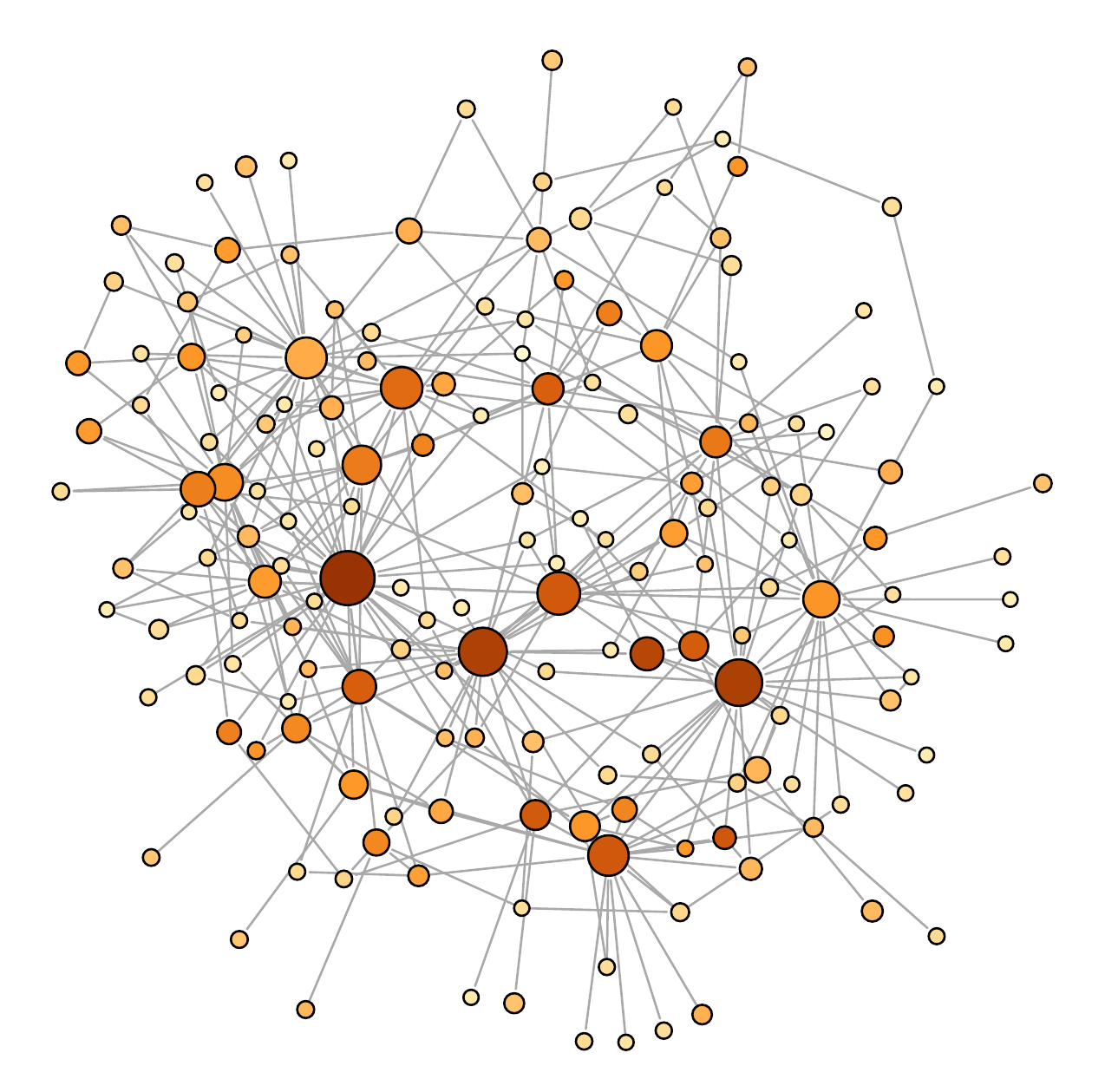}
\hspace{0.005\textwidth}
\includegraphics[width=0.49\textwidth]{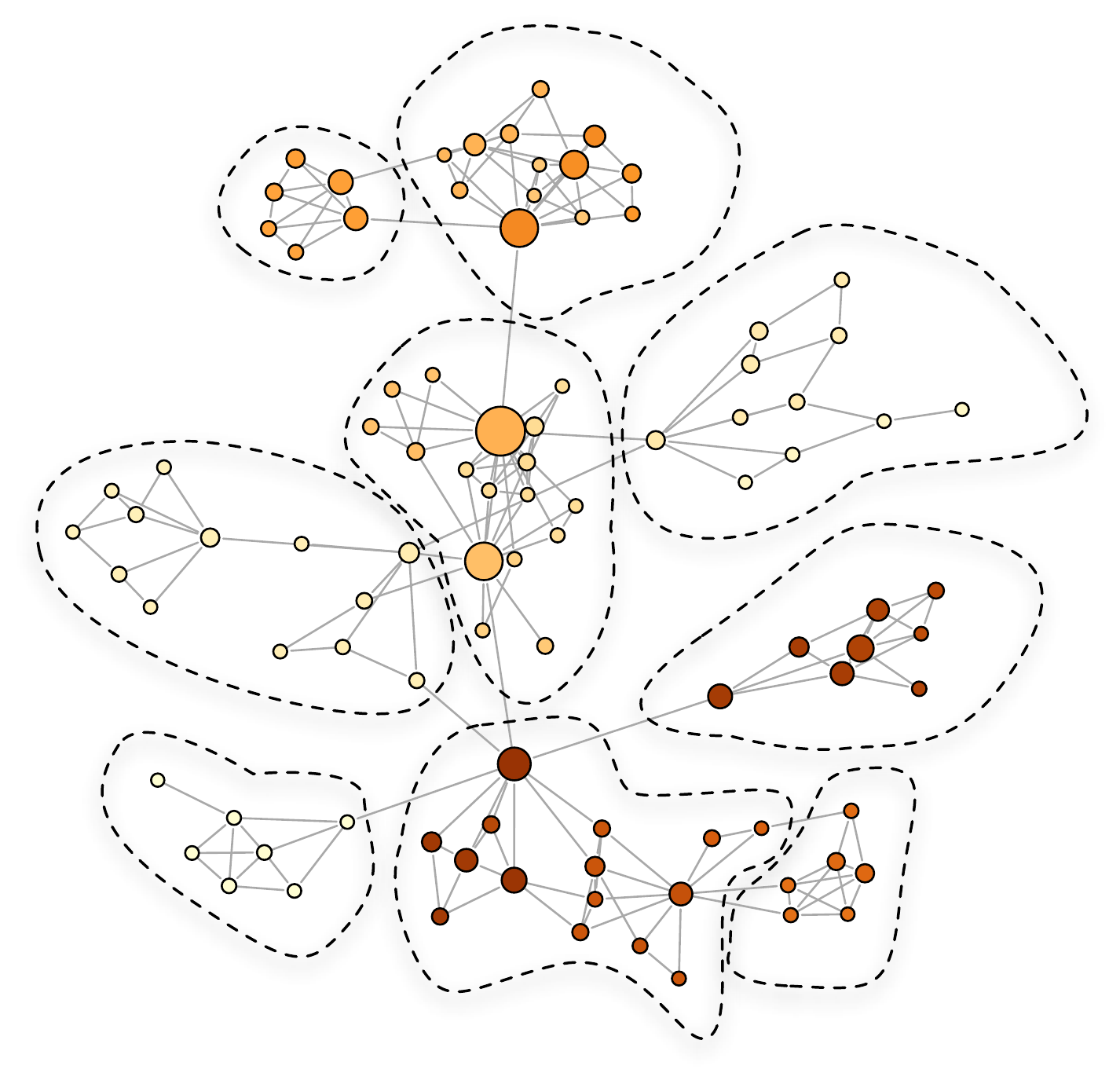}
\caption{Visualization of network communities in QAP instances. {\bf Left}: a uniform instance of size 9. {\bf Right}: a real-like instance of size. Node size is proportional to basin size, and node color to fitness (the darker, the better). Communities are highlighted in the right picture, which was not possible in the left one.\label{fig:qap-comm}}
\end{center}
\end{figure}

The real-like instances have significantly more cluster structure than the uniform instances. This can be appreciated visually in Fig.~\ref{fig:qap-comm} illustrating the community structures of two particular instances. These two selected cases have the highest $Q$ values of their respective classes, but they represent a general trend. For the real-like instance (Fig.~\ref{fig:qap-comm}, right) the groups of local optima are recognizable and form well separated clusters (encircled with dotted lines), which is also reflected in the high corresponding modularity value $Q=0.79$. In contrast, the the LON of the uniform instance  (Fig.~\ref{fig:qap-comm}, left)  has some modularity, with a $Q=0.53$, but the communities are hard to represent graphically, and thus are not shown in the picture.

The LONs community structure is likely to have consequences on the heuristic algorithms used to search the corresponding landscapes. According  to the level of modularity, different search strategies would be more efficient. For example, we can envision that for real-like instances, a local search algorithm may require stronger perturbation mechanisms to escape a cluster of local optima with poor quality solutions.

\section{Conclusions}

A network-based model of combinatorial landscapes is described and a thorough analysis is presented for two example combinatorial landscapes: $NK$ landscapes and the Quadratic Assignment Problem. A network model requires defining its nodes and edges. Nodes are the landscape local optima obtained with a best-improvement hill-climbing algorithm; edges are defined in two alternative ways: one is based on the transition probabilities between basins of attraction, the other on the transition (escape) probabilities from local optima. The model, therefore, compresses the  fitness landscapes into a more manageable mathematical object. New features can be measured in this model, coming from the science of complex networks  such as the degree distribution, clustering coefficient, shortest path length, disparity and community structure. Results from the studied landscapes show that local optima networks share some features with complex networks: basin sizes are not uniformly distributed, weight distributions are not normal,  path lengths to the global optimum can be short, clustering coefficients can be high, and networks can have community structure.

The results clearly show that the search difficulty on the studied landscapes, which may be either known a priori or empirically estimated, correlate with some fundamental LON features such as the number of nodes, size of basins, shortest path length to the global optimum, out-degree, disparity, and degree of assortativity. Indeed, some of these metrics were used successfully to construct a statistical predictive model of search performance. The network analysis also revealed interesting topological differences on the distribution of local optima for different classes of problem instances. These differences may lead to designing search heuristics that can adapt and thus exploit the landscape structure.

Local optima networks can be seen as a generic model of combinatorial landscapes based on defining convergence points of simple heuristics in the search space. In this work, the convergence points are local optima, and edges are transition probabilities between these points which also reflect the path of a local search. From a mathematical point of view, LONs reduce the study of the whole transition matrix between solutions of a local search by a smaller transition matrix between local optima. The search dynamic is then decomposed into two time scales: one to reach local optima,  the other to traverse between local optima.  From LON graphs and their corresponding transition matrices, it should be possible to conduct a Markov chain analysis and thus compute running times or expected performance.

Another research direction is to use LONs for automated parameter tuning and design of heuristic search methods. Some network metrics can be estimated without knowing the global optimum beforehand. These metrics coupled with adequate performance prediction models open up exciting possibilities. Our current analysis requires the exhaustive enumeration of the search space; with standard sampling methods, larger search spaces could be studied.  We plan to continue working on the afore mentioned directions and extend this analysis to other combinatorial optimization problems.

\bibliographystyle{plain}
\small

\begin{thebibliography}{}

\end{thebibliography}


\begin{thebibliography}{10}

\bibitem{auger2005performance}
A.~Auger and N.~Hansen.
\newblock Performance evaluation of an advanced local search evolutionary
  algorithm.
\newblock In {\em Proceedings of the IEEE Congress on Evolutionary Computation,
  CEC 2005}, volume~2, pages 1777--1784. IEEE, 2005.

\bibitem{barabasi99}
A-L. Barabasi and R.~Albert.
\newblock Emergence of scaling in random networks.
\newblock {\em Science}, 286:509--512, 1999.

\bibitem{barnett98}
L.~Barnett.
\newblock Ruggedness and neutrality - the {NK}p family of fitness landscapes.
\newblock In C.~Adami, R.~K. Belew, H.~Kitano, and C.~Taylor, editors, {\em
  Proceedings of the Sixth International Conference on Artificial Life, ALIFE
  VI}, pages 18--27. ALIFE, The MIT Press, 1998.

\bibitem{bart05}
M.~Barth\'elemy, A.~Barrat, R.~Pastor-Satorras, and A.~Vespignani.
\newblock Characterization and modeling of weighted networks.
\newblock {\em Physica A}, 346:34--43, 2005.

\bibitem{Battiti2009}
R.~Battiti, M.~Brunato, and F.~Mascia.
\newblock {\em Reactive Search and Intelligent Optimization}, volume~45 of {\em
  Operations Research/Computer Science Interfaces Series}.
\newblock Springer, 2009.

\bibitem{boese94}
K.~D. Boese, A.~B. Kahng, and S.~Muddu.
\newblock A new adaptive multi-start technique for combinatorial global
  optimizations.
\newblock {\em Operations Research Letters}, 16:101--113, 1994.

\bibitem{chicano2012}
F.~Chicano, F.~Daolio, G.~Ochoa, S.~Verel, M.~Tomassini, and E.~Alba.
\newblock Local optima networks, landscape autocorrelation and heuristic search
  performance.
\newblock In {\em Proceedings of Parallel Problem Solving from Nature - PPSN
  XII}, volume 7492 of {\em Lecture Notes in Computer Science}, pages 337--347.
  Springer, 2012.

\bibitem{Clauset2009}
A.~Clauset, C.~Shalizi, and M.~Newman.
\newblock Power-law distributions in empirical data.
\newblock {\em SIAM Review}, 51(4):661--703, 2009.

\bibitem{daolio2011communities}
F.~Daolio, M.~Tomassini, S.~Verel, and G.~Ochoa.
\newblock Communities of minima in local optima networks of combinatorial
  spaces.
\newblock {\em Physica A: Statistical Mechanics and its Applications},
  390(9):1684--1694, 2011.

\bibitem{cec10}
F.~Daolio, S.~Verel, G.~Ochoa, and M.~Tomassini.
\newblock Local optima networks of the quadratic assignment problem.
\newblock In {\em Proceedings of the IEEE Congress on Evolutionary Computation,
  CEC 2010}, pages 1--8. IEEE Press, 2010.

\bibitem{daolio2012local}
F.~Daolio, S.~Verel, G.~Ochoa, and M.~Tomassini.
\newblock Local optima networks and the performance of iterated local search.
\newblock In {\em Proceedings of the Genetic and Evolutionary Computation
  Conference, GECCO 2012}, pages 369--376. ACM, 2012.

\bibitem{dorogotsev}
S.~N. Dorogotsev.
\newblock {\em Lectures on Complex Networks}.
\newblock Oxford University Press, Oxford, UK, 2010.

\bibitem{doye02}
J.~P.~K. Doye.
\newblock The network topology of a potential energy landscape: a static
  scale-free network.
\newblock {\em Physical Review Letter}, 88:238701, 2002.

\bibitem{doye05}
J.~P.~K. Doye and C.~P. Massen.
\newblock Characterizing the network topology of the energy landscapes of
  atomic clusters.
\newblock {\em Jourla of Chemical Physics}, 122:084105, 2005.

\bibitem{santo1}
S.~Fortunato.
\newblock Community detection in graphs.
\newblock {\em Physics Reports}, 486:75--174, 2010.

\bibitem{garnier01}
J.~Garnier and L.~Kallel.
\newblock Efficiency of local search with multiple local optima.
\newblock {\em SIAM Journal on Discrete Mathematics}, 15(1):122--141, 2001.

\bibitem{gfeller-07}
D.~Gfeller, P.~De~Los Rios, A.~Caflisch, and F.~Rao.
\newblock Complex network analysis of free-energy landscapes.
\newblock {\em Proc. Nat. Acad. Sci. USA}, 104(6):1817--1822, 2007.

\bibitem{HainsWH11}
D.~Hains, L.~D. Whitley, and A.~E. Howe.
\newblock Revisiting the big valley search space structure in the {TSP}.
\newblock {\em Journal of the Operational Research Society}, 62(2):305--312,
  2011.

\bibitem{jones95b}
T.~Jones.
\newblock {\em Evolutionary Algorithms, Fitness Landscapes and Search}.
\newblock PhD thesis, University of New Mexico, Albuquerque, 1995.

\bibitem{Kauffman1987}
S.~Kauffman and S.~Levin.
\newblock Towards a general theory of adaptive walks on rugged landscapes.
\newblock {\em Journal of Theoretical Biology}, 128:11--45, 1987.

\bibitem{kauffman93}
S.~A. Kauffman.
\newblock {\em The Origins of Order}.
\newblock Oxford University Press, New York, 1993.

\bibitem{kaul06}
H.~Kaul and S.~H. Jacobson.
\newblock New global optima results for the {K}auffman {$NK$} model: handling
  dependency.
\newblock {\em Mathematical Programming}, 108(2-3, Ser. B):475--494, 2006.

\bibitem{Knowles2003emo}
J.~Knowles and D.~Corne.
\newblock Instance generators and test suites for the multiobjective quadratic
  assignment problem.
\newblock In {\em Proceedings of the Evolutionary Multi-Criterion Optimization
  Conference (EMO 2003)}, number 2632 in LNCS, pages 295--310. Springer, 2003.

\bibitem{Koopmans57}
T.~C. Koopmans and M.~Beckmann.
\newblock Assignment problems and the location of economic activities.
\newblock {\em Econometrica}, 25(1):53--76, 1957.

\bibitem{limic04}
V.~Limic and R.~Pemantle.
\newblock More rigorous results on the kauffman-levin model of evolution.
\newblock {\em Annals of Probability}, 32:2149, 2004.

\bibitem{lourenco:2002}
H.~R. Louren{\c c}o, O.~Martin, and T.~St{\" u}tzle.
\newblock Iterated local search.
\newblock In {\em Handbook of Metaheuristics}, volume~57 of {\em International
  Series in Operations Research \& Management Science}, pages 321--353. Kluwer
  Academic Publishers, 2002.

\bibitem{doye05-comm}
C.~P. Massen and J.~P.~K. Doye.
\newblock Identifying communities within energy landscapes.
\newblock {\em Physical Review E}, 71:046101, 2005.

\bibitem{merz98}
P.~Merz and B.~Freisleben.
\newblock Memetic algorithms and the fitness landscape of the graph
  bi-partitioning problem.
\newblock In {\em Proceedings of Parallel Problem Solving from Nature, PPSN V},
  volume 1498 of {\em Lecture Notes in Computer Science}, pages 765--774.
  Springer-Verlag, 1998.

\bibitem{newman98}
M.~Newman and R.~Engelhardt.
\newblock Effect of neutral selection on the evolution of molecular species.
\newblock {\em Proc. Royal Society London B.}, 256:1333--1338, 1998.

\bibitem{newman03}
M.~E.~J. Newman.
\newblock The structure and function of complex networks.
\newblock {\em SIAM Review}, 45:167--256, 2003.

\bibitem{newman-book}
M.~E.~J. Newman.
\newblock {\em Networks: An Introduction}.
\newblock Oxford University Press, Oxford, UK, 2010.

\bibitem{gecco08}
G.~Ochoa, M.~Tomassini, S.~Verel, and C.~Darabos.
\newblock A study of {NK} landscapes' basins and local optima networks.
\newblock In {\em Proceedings of the Genetic and Evolutionary Computation
  Conference, GECCO 2008}, pages 555--562. ACM, 2008.

\bibitem{ppsn10}
G.~Ochoa, S.~Verel, and M.~Tomassini.
\newblock First-improvement vs. best-improvement local optima networks of nk
  landscapes.
\newblock In {\em Proceedings of Parallel Problem Solving from Nature - PPSN
  XI}, volume 6238 of {\em Lecture Notes in Computer Science}, pages 104--113.
  Springer, 2010.

\bibitem{rao-caflisch04}
F.~Rao and A.~Caflisch.
\newblock The network topology of a potential energy landscape: a static
  scale-free network.
\newblock {\em Journal of Molecular Biology}, 342:299--306, 2004.

\bibitem{reeves99}
C.~R. Reeves.
\newblock Landscapes, operators and heuristic search.
\newblock {\em Annals of Operations Research}, 86:473--490, 1999.

\bibitem{reidys2002combinatorial}
C.M. Reidys and P.F. Stadler.
\newblock {Combinatorial landscapes}.
\newblock {\em SIAM review}, 44(1):3--54, 2002.

\bibitem{ROS:96}
H.~Ros{\'e}, W.~Ebeling, and T.~Asselmeyer.
\newblock The density of states - a measure of the difficulty of optimisation
  problems.
\newblock In H.-M.~Voigt et~al., editor, {\em Parallel Problem Solving from
  Nature, {PPSN IV}}, volume 1141 of {\em Lecture Notes in Computer Science},
  pages 208--217. Springer, Berlin, Heidelberg, New York, 1996.

\bibitem{sahni76}
S.~Sahni and T.~Gonzalez.
\newblock {P-complete approximation problems}.
\newblock {\em Journal of the ACM (JACM)}, 23(3):555--565, 1976.

\bibitem{stillinger95}
F.H. Stillinger.
\newblock A topographic view of supercooled liquids and glass formation.
\newblock {\em Science}, 267:1935--1939, 1995.

\bibitem{Taillard1995}
\'{E}.~D. Taillard.
\newblock Comparison of iterative searches for the quadratic assignment
  problem.
\newblock {\em Location Science}, 3(2):87 -- 105, 1995.

\bibitem{pre08}
M.~Tomassini, S.~Verel, and G.~Ochoa.
\newblock Complex-network analysis of combinatorial spaces: The {NK} landscape
  case.
\newblock {\em Physical Revview E}, 78(6):066114, 2008.

\bibitem{nsc06}
L.~Vanneschi, M.~Tomassini, P.~Collard, and S.~Verel.
\newblock Negative slope coefficient. a measure to characterize genetic
  programming fitness landscapes.
\newblock In {P. Collet et al.}, editor, {\em Proceedings of the 9th European
  Conference on Genetic Programming}, volume 3905 of {\em Lecture Notes in
  Computer Science}, pages 178--189. Springer, Berlin, Heidelberg, New York,
  2006.

\bibitem{ea11}
S.~Verel, F.~Daolio, G.~Ochoa, and M.~Tomassini.
\newblock Local optima networks with escape edges.
\newblock In {\em Proceedings of the International Conference on Artificial
  Evolution, EA-2011}, volume 7401 of {\em Lecture Notes in Computer Science},
  pages 49--60. Springer, 2012.

\bibitem{alife08}
S.~Verel, G.~Ochoa, and M.~Tomassini.
\newblock The connectivity of {NK} landscapes' basins: a network analysis.
\newblock In {\em Proceedings of the Eleventh International Conference on
  Artificial Life, ALIFE XI}, pages 648--655. MIT Press, Cambridge, MA, 2008.

\bibitem{tec11}
S.~Verel, G.~Ochoa, and M.~Tomassini.
\newblock Local optima networks of {NK} landscapes with neutrality.
\newblock {\em IEEE Transactions on Evolutionary Computation}, 15(6):783--797,
  2011.

\bibitem{watts-strogatz-98}
D.~J. Watts and S.~H. Strogatz.
\newblock Collective dynamics of 'small-world' networks.
\newblock {\em Nature}, 393:440--442, 1998.

\bibitem{weinberger90}
E.~D. Weinberger.
\newblock Correlated and uncorrelated fitness landscapes and how to tell the
  difference.
\newblock {\em Biological Cybernetics}, 63:325--336, 1990.

\bibitem{weinberger91}
E.~D. Weinberger.
\newblock Local properties of {K}auffman's {NK} model, a tuneably rugged energy
  landscape.
\newblock {\em Phys. Rev. A}, 44:6399--6413, 1991.

\bibitem{WhitleyHH10}
D.~Whitley, D.~Hains, and A.~E. Howe.
\newblock A hybrid genetic algorithm for the traveling salesman problem using
  generalized partition crossover.
\newblock In {\em Proceedings of Parallel Problem Solving from Nature, {PPSN
  XI}}, volume 6238 of {\em Lecture Notes in Computer Science}, pages 566--575.
  Springer, 2010.

\bibitem{Wright:32}
S.~Wright.
\newblock The roles of mutation, inbreeding, crossbreeding and selection in
  evolution.
\newblock In D.~F. Jones, editor, {\em Proceedings of the Sixth International
  Congress on Genetics}, volume~1, pages 356--366, 1932.

\end{thebibliography}

\end{document}